\documentclass[final,5p,times,twocolumn]{elsarticle}

\usepackage[utf8]{inputenc}
\usepackage{amsmath,amsfonts}
\usepackage{array}
\usepackage[caption=false,font=normalsize,labelfont=sf,textfont=sf]{subfig}
\usepackage{textcomp}
\usepackage{stfloats}
\usepackage{url}
\usepackage{verbatim}
\usepackage{graphicx}
\usepackage{threeparttable}
\usepackage{multirow}
\usepackage{stfloats}
\usepackage{amssymb}
\usepackage{color}
\usepackage[ruled,vlined]{algorithm2e}
\usepackage{setspace}
\usepackage{booktabs}
\usepackage{multirow}
\usepackage[normalem]{ulem}
\useunder{\uline}{\ul}{}

\usepackage[table]{xcolor}
\definecolor{lightred}{RGB}{255, 220, 220}
\definecolor{lightgreen}{RGB}{220, 255, 220}
\definecolor{lightyellow}{RGB}{255, 250, 205}

\makeatletter
\renewcommand{\maketag@@@}[1]{\hbox{\m@th\normalsize\normalfont#1}}%
\makeatother

\usepackage{tikz,xcolor,hyperref}

\definecolor{lime}{HTML}{A6CE39}
\DeclareRobustCommand{\orcidicon}{
\begin{tikzpicture}
\draw[lime, fill=lime] (0,0)
circle[radius=0.16]
node[white]{{\fontfamily{qag}\selectfont \tiny \.{I}D}}; 
\end{tikzpicture}
\hspace{-2mm}
}
\foreach \x in {A, ..., Z}{
\expandafter\xdef\csname orcid\x\endcsname{\noexpand\href{https://orcid.org/\csname orcidauthor\x\endcsname}{\noexpand\orcidicon}}
}

\journal{Computer-Aided Civil and Infrastructure Engineering}

\begin{document}

\begin{frontmatter}


\title{
AdvSerial: Physical Adversarial Attacks on Infrastructure-mounted Pedestrian Detectors via Semantic Feature Suppression
}


\author[label1,label2]{Yuanhao Huang}
\ead{yuanhao\_huang@buaa.edu.cn}

\author[label1,label3]{Yilong Ren\corref{cor1}}
\ead{yilongren@buaa.edu.cn}

\author[label4]{Jinlei Wang}
\ead{20231800822@imut.edu.cn}


\author[label1,label2]{Xuesong Bai}
\ead{xs\_bai@buaa.edu.cn}

\author[label1]{Jinchuan Zhang}
\ead{22376270@buaa.edu.cn}

\author[label1,label3]{Haiyang Yu\corref{cor1}}
\ead{hyyu@buaa.edu.cn}


\cortext[cor1]{Corresponding author}

\cortext[cor2]{Funding: This work was supported by the Category-A of Excellent Research Groups Project (T2588101) and fundamental research funds for the central universities.}

\affiliation[label1]{organization={School of Transportation Science and Engineering, Beihang University},
            postcode={100191}, 
            state={Beijing},
            country={P.R China}}

\affiliation[label2]{organization={State Key Lab of Intelligent Transportation System},
            postcode={100191}, 
            state={Beijing},
            country={P.R China}}
            
\affiliation[label3]{organization={Zhongguancun Laboratory},
            postcode={100191}, 
            state={Beijing},
            country={P.R China}}
            
\affiliation[label4]{organization={School of Systems Science and Engineering, Sun Yat-Sen University},
            postcode={510006}, 
            state={Guangzhou},
            country={P.R China}}

\begin{abstract}
AI-based visual perception systems are increasingly deployed in infrastructure surveillance, including roadside monitoring units, highway cameras, and smart-city pedestrian management systems. The security vulnerability of these systems to physical adversarial attacks poses a direct threat to the reliable operation of transportation infrastructure.  We propose AdvSerial, a dynamic 2D--3D joint optimization framework for generating continuous high-angle physical adversarial patches against pedestrian detectors in infrastructure-based scenarios. We UV-map a boundary-aware quilted texture onto 3D garments, combine 2D digital attacks with 3D sparse- and continuous-frame rendering, and explicitly suppress person-specific semantic features while enforcing temporal continuity. A Feature Smooth Quilting strategy reduces visible patch boundaries and bounds cross-seam feature discontinuities. A serial-frame loss encourages long uninterrupted sequences of detection failures. In physical world experiments, AdvSerial achieves a $74.8\%$ attack success rate on YOLO-v5 and degrades mean detection confidence from $84.30\%$ to $39.38\%$. Experiments spanning eight detectors with different architectures demonstrate strong transferability. Notably, it achieves an $89.71\%$ attack success rate on YOLO-v2 and resists both patch-detection defenses (NapGuard) and 3D-temporal perception (Sparse4D-v3). The results reveal persistent, temporally consistent failure modes under high-angle surveillance, and motivate the design of motion-aware and 3D-aware defenses for security-critical infrastructure deployments.
\end{abstract}



\end{frontmatter}



\section{Introduction}
\label{Introduction}

AI-based visual perception has become integral to infrastructure surveillance and intelligent transportation. Pedestrian detection now underpins safety-critical functions ranging from highway monitoring to smart-city crowd management. However, deep neural network based detectors are vulnerable to adversarial perturbations. Physically realizable patches can severely compromise recognition accuracy and system reliability in deployed environments. Infrastructure-mounted cameras typically operate from elevated viewpoints. High-angle surveillance therefore introduces geometric distortion, reduced patch visibility, and non-rigid clothing deformation. These factors collectively challenge both detection robustness and the design of adversarial security assessments~\cite{hu2025exploring}.

Existing physical adversarial attacks have largely been designed for frontal, street-level views. AdvTshirt maps perturbations onto the front of a shirt~\cite{xu2020adversarial}, and T-SEA overlays patches onto pedestrian images from the INRIA dataset~\cite{huang2023t}. Both rely on simplified 2D transformations that frequently fail under real-world lighting, viewpoint, and posture variations~\cite{huang2026advreal}. Meanwhile, attack success rates achieved in digital simulations do not reliably predict physical-world performance~\cite{hingun2023reap}. To narrow this gap, recent work employs differentiable 3D rendering for more realistic adversarial training. DTA applies differentiable transformation attacks via a neural renderer~\cite{suryanto2022dta}. AdvReal models cloth wrinkles and lighting through PyTorch3D~\cite{huang2026advreal}. DynamicPAE explicitly aligns distribution differences between training and deployment environments~\cite{hu2025dynamicpae}. Nevertheless, adversarial patches trained on sparse views still struggle to sustain detection failures across continuous video sequences. Their actual impact on deployed surveillance systems remains significantly lower than expected~\cite{dong2025feature}.

In practice, infrastructure surveillance systems rely on detection-based pipelines as their primary perception mechanism, where continuous pedestrian detection feeds directly into downstream tracking and behavioral analysis~\cite{gu2026assessing}. Two key gaps persist. First, existing attacks target discrete views or single frames and lack systematic designs that remain effective across long video sequences, multiple poses, and steep overhead angles. Sustaining detection failures across consecutive frames is therefore the precise condition required to compromise real-world surveillance operations. Second, no mechanism-oriented analysis has explained how such attacks suppress person-specific features in modern detection architectures. The failure modes of infrastructure-deployed detectors therefore remain poorly understood.

To address these gaps, we propose \textbf{AdvSerial}, a dynamic 2D--3D joint optimization framework for generating continuous, high-angle physical adversarial patches against pedestrian detectors in infrastructure-based scenarios. We UV-map seamless textured patches onto 3D garments, combine 2D digital attacks with 3D sparse- and continuous-frame rendering, and explicitly suppress person-specific semantic features while enforcing temporal continuity. To the best of our knowledge, this is the first high-angle physical adversarial attack and security assessment framework specifically designed for infrastructure-mounted pedestrian detection systems, addressing a critical but underexplored vulnerability in AI-enabled civil infrastructure. The specific contributions are as follows.

\begin{itemize}
\item[1)] We propose AdvSerial, a dynamic 2D--3D joint optimization framework for physical adversarial patches targeting pedestrian detectors in infrastructure-based high-angle settings. We introduce Feature Smooth Quilting, a feature-sensitivity weighted quilting strategy that routes tile seams through detector-insensitive regions. This strategy mathematically bounds cross-seam feature discontinuities while preserving adversarial potency.

\item[2)] We design a Serial Frame Loss (SFL) that couples semantic feature suppression with temporal continuity. SFL steers optimization toward flat, temporally shared minima, suppressing adversarial flickering and inducing durable confidence reduction across varying pose, scale, illumination, and background.

\item[3)] We conduct a comprehensive evaluation across frame-based detectors and sequence-based trackers in both simulation and the physical world, including sequence-level tracking and 3D-temporal defense pipelines. We further analyze the attack mechanism via t-SNE and Grad-CAM, showing that AdvSerial achieves stronger robustness, transferability, and stealth than existing methods. These results expose temporally consistent failure modes in infrastructure surveillance and provide actionable guidance for the design of motion-aware and 3D-aware defenses.
\end{itemize}

The remainder of this paper is organized as follows. Section~\ref{related_works} reviews related work on physical adversarial attacks from multiple angles and the mechanisms of semantic feature suppression. Section~\ref{method} presents the proposed AdvSerial framework, including the 2D--3D joint attack structure, Feature Smooth Quilting, pose generation, and the Serial Frame Loss. Section~\ref{experiments} describes the experimental setup and reports results covering SOTA comparisons, cross-model transferability, tracking pipeline evaluation, environmental and pitch-angle robustness, defense robustness, and physical-world validation. Section~\ref{discussion} analyzes the attack mechanism and discusses limitations and future directions. Section~\ref{conclusion} concludes the paper.

\section{Related Works}
\label{related_works}

\subsection{Adversarial Attacks from Multiple Angles}
Even minor shifts in camera angle or object pose can cause a sharp drop in ASR of physical adversarial patches. To improve robustness under realistic deployment conditions, researchers have turned attention to multi-angle optimization strategies. AdvTexture applies a scalable patch over the entire garment surface to improve pose- and view-robustness~\cite{hu2022adversarial}, while AdvReal further models cloth wrinkles and lighting to better match real-world conditions~\cite{huang2026advreal}. Beyond localized patches, recent works have explored full-body coatings for 3D objects. For instance, coated adversarial camouflages (CAC) proposed a 3D rendering framework and dense proposal attacks to maintain effectiveness across arbitrary viewpoints~\cite{ijcai2022p125}. DynamicPAE enhances the generalization ability of physical attacks in dynamic and variable scenarios by explicitly modeling and aligning the distribution differences between the attack environment and real observation conditions during the training phase~\cite{hu2025dynamicpae}.

A common requirement of these methods is end-to-end differentiability through the rendering pipeline. Differentiable renderers such as SoftRas~\cite{liu2019softras} and PyTorch3D~\cite{ravi2020pytorch3d} enable gradient-based texture optimization by replacing discrete rasterization with soft-blending formulations that provide locally smooth gradients. However, the smoothness guarantee of these renderers is confined to individual mesh faces; when adversarial textures are tiled across the garment surface, feature continuity at tile seams requires additional mechanisms. Recent work has begun to address gradient-level challenges in 3D adversarial texture optimization. Liang~et~al.~\cite{liang2025gradient} proposed gradient calibration to resolve inconsistent sampling densities across distances, while GRAC~\cite{liang2025grac} introduced gradient reweighting to eliminate conflicts in multi-angle updates. These methods operate on the texture optimization process itself but do not address the seam discontinuities that arise when adversarial tiles are quilted onto clothing surfaces. In contrast, infrastructures frequently rely on high-mounted cameras that capture pedestrians from elevated angles. Such views introduce distortion, reduced patch visibility, and inconsistent spatial context, presenting new challenges for adversarial patch design in object detectors~\cite{zhang2023boosting}. A critical gap is the existing adversarial patches fail to maintain robust adversarial effects under dynamic high-angle conditions. We must bridge the gap between static texture perturbations and infrastructure-specific geometric constraints, which is essential for evaluating the actual security of infrastructure detectors.

\subsection{Mechanisms of Semantic Features Suppression}
\label{Semantic_Feature}

A fundamental question behind adversarial patch design is which visual features the target detector relies on for person recognition. Geirhos et al.~\cite{geirhos2018imagenet} proposed that ImageNet-trained CNNs are biased towards texture, a view that historically motivated many attacks to concentrate on texture manipulation. However, Burgert et al.~\cite{burgert2026imagenet} recently demonstrated through controlled suppression experiments that CNNs depend predominantly on local shape features, with this reliance shifting across training regimes and domains. From this mechanistic perspective, related works divide into two families. The first adopts \textbf{feature activation guidance}, designing patches that redirect model attention through localized discriminative patterns that hijack feature maps and logits~\cite{huang2026advreal, deng2025targeted}, producing spurious detections rather than complete suppression. The second emphasizes \textbf{semantic feature suppression}, weakening person-specific feature responses across the person's spatial extent~\cite{hu2022adversarial}. Aligning with Burgert et al.'s findings, suppression-oriented attacks dismantle the continuous local shape structures that modern detectors fundamentally rely on, directly targeting the detection decision margin and yielding more stable confidence reduction across views and time.

Spatial smoothness plays a critical role in the effectiveness and transferability of adversarial perturbations. Dabouei et al.~\cite{dabouei2020smoothfool} demonstrated that smooth, low-frequency perturbations exhibit superior cross-model transferability and defense robustness by avoiding overfitting to model-specific high-frequency decision boundaries. From a theoretical perspective, Gatys et al.~\cite{gatys2015texture} established that texture identity resides in deep feature statistics rather than raw pixel values, while Khromov and Singh~\cite{khromov2024lipschitz} showed that the spatially non-uniform Lipschitz constants of deep networks imply that pixel-level smoothness at tile boundaries does not guarantee feature-level continuity. Consequently, high-contrast seams between adjacent tiles introduce precisely the high-frequency artifacts that degrade transferability and trigger anomaly-based defenses.

However, most existing patch designs overlook appearance continuity at patch boundaries. Simple tiling or overlay operations introduce sharp, unnatural edges that disrupt the lighting and geometric consistency expected in real scenes, especially under oblique views~\cite{hu2022adversarial}. While methods like CAC achieve multi-view robustness through 3D surface mapping, they often overlook texture seam artifacts during 2D-to-3D projection and temporal flickering in continuous video streams. The tension between strong digital perturbations and physically realistic seamless integration therefore remains a key obstacle to achieving robust multi-angle suppression attacks in infrastructure surveillance scenarios.

\section{Methodology}
\label{method}

\begin{figure*}[t!]
\centering
\includegraphics[width=14cm]{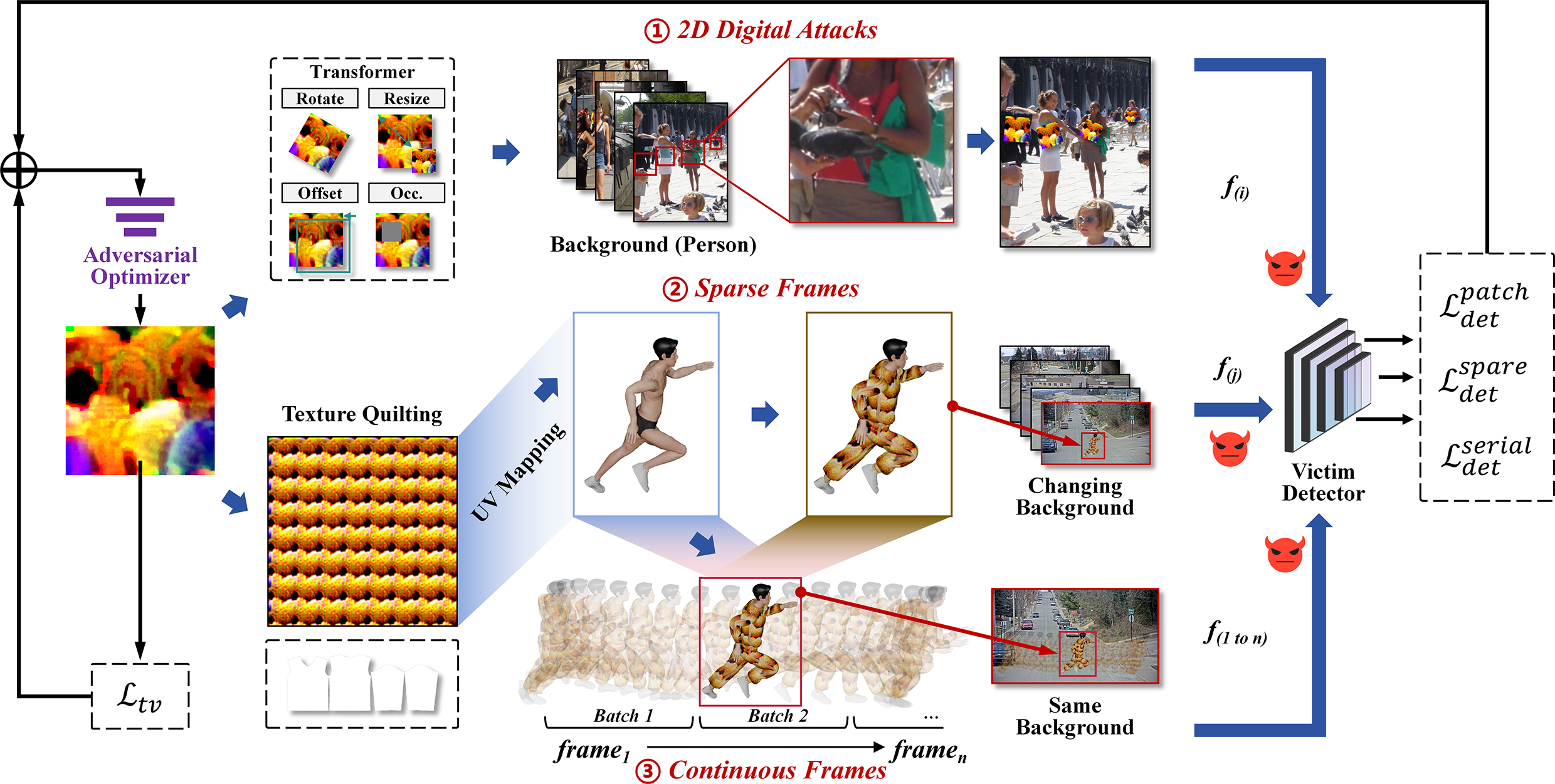}
\caption{Overview of the proposed adversarial attack method. The pipeline consists of three joint optimization branches: 2D digital attacks, 3D sparse-frame rendering, and 3D continuous-frame rendering with Serial Frame Loss.}
\label{structure}
\vspace{-4 mm} 
\end{figure*}

In this section, the research objective is first clarified and the optimization problem is formally defined. The proposed method is then presented in detail, including the adversarial attacks framework, the quilting-based optimization strategy, and the design of the continuous-frame adversarial loss.

\subsection{Problem Definition}

This study investigates the generation of physically realizable adversarial patches that can mislead pedestrian detectors in infrastructure-based surveillance. We focus on robustness under elevated views and temporal consistency across video sequences.

Let $P(X, Y)$ be the joint distribution of pedestrian image data, where $X \subseteq \mathbb{R}^{H\times W\times 3}$ is the image space and $Y \subseteq \{1, \ldots, C\}$ the label space. A pretrained detector $f_\theta: X \to [0, 1]^C$ maps an image $x \in X$ to a set of bounding boxes with class-confidence scores in $[0,1]^C$. Given samples $\{(x_1, y_1), \ldots, (x_n, y_n)\} \sim P(X, Y)$, we seek a patch pattern $\delta \in \mathcal{D}$ such that the rendered adversarial image $x_i^{\text{adv}} = \mathcal{R}(x_i, \delta, \phi_i)$, with transformation parameters $\phi_i$, yields consistent detection failures across conditions.

The optimization problem is
\begin{equation}
\delta^* = \arg\min_{\delta} \; \frac{1}{n} \sum_{i=1}^{n} \mathcal{L}(f_\theta(x_i^{\text{adv}}), y_i) + \lambda \mathcal{R}_{\text{total}}(\delta),
\end{equation}
where $\mathcal{L}$ is the adversarial loss that suppresses detection confidence or localization accuracy, and $\mathcal{R}_{\text{total}}(\delta)$ combines spatial smoothness, view robustness, and temporal continuity. The weighting factor $\lambda$ balances attack strength against physical plausibility and temporal stability.

\subsection{Adversarial Attack Structure}
\label{adv_structure}

To achieve the above optimization goals, we designed a 2D–3D joint adversarial attack framework. The framework achieves adversarial optimization by simulating and rendering actual deployment scenarios for the continuous detection of pedestrians at multiple angles and scales. As shown in Fig.~\ref{structure}, our training pipeline consists of three parallel branches as follows. (i) a 2D digital attack branch on pedestrian datasets, (ii) a 3D sparse-frame branch with randomly sampled poses and views, and (iii) a 3D continuous-frame branch for temporal consistency. The gradients from all branches are jointly backpropagated to optimize the same learnable texture $\theta$.

\subsubsection{\textbf{2D Digital Attacks}}
\label{structure_2d}

In the 2D branch, we overlay a square patch covering approximately $\alpha\%$ of the visible torso area, where $\alpha$ is computed from the size of the person’s bounding box. The patch location is offset around the upper body with random translation and rotation, and standard photometric augmentations (brightness, contrast, Gaussian noise) are applied to enhance robustness against intra-class appearance variations. The resulting images are used to attack the victim detector and compute the detection loss. This component is designed to guide rapid gradient descent and improve the generalization performance of adversarial attacks within pedestrian classes.

\subsubsection{\textbf{Sparse Frames}}

To enhance spatial robustness under varying views and environmental conditions, the sparse frame branch employs 3D-rendered human models with randomly sampled poses, camera positions, and background scenes. This process simulates diverse views, scales, and body geometries, thereby improving attack reliability across heterogeneous observation angles. In particular, the models can simulate introduced by clothing folds and curved body surfaces. The use of quilting-based patch construction further mitigates boundary artifacts and improves generalization against patch detection and defense methods.

\subsubsection{\textbf{Continuous Frames}}

In contrast to sparse frames, continuous frames generate adversarial samples by rendering a sequence of action frames over time. All frames share the same adversarial texture from the current training epoch, while their shapes and occlusions vary with changes in body posture. The sequence is divided into $n$ fixed-size batches, which are merged and input to the victim detector to compute the serial loss.

The optimization strategy jointly updates the adversarial patch using feedback from all three branches. The texture is parameterized as a learnable pattern and applied to 3D garments through the quilting and pose generation process introduced in Sec.~\ref{quilting} and Sec.~\ref{poseture}. The serial loss, along with other detection and regularization terms, is detailed in Sec.~\ref{loss}.


\subsection{Feature Smooth Quilting}
\label{quilting}

\begin{figure}[t!]
\centering
\includegraphics[width=8.6 cm]{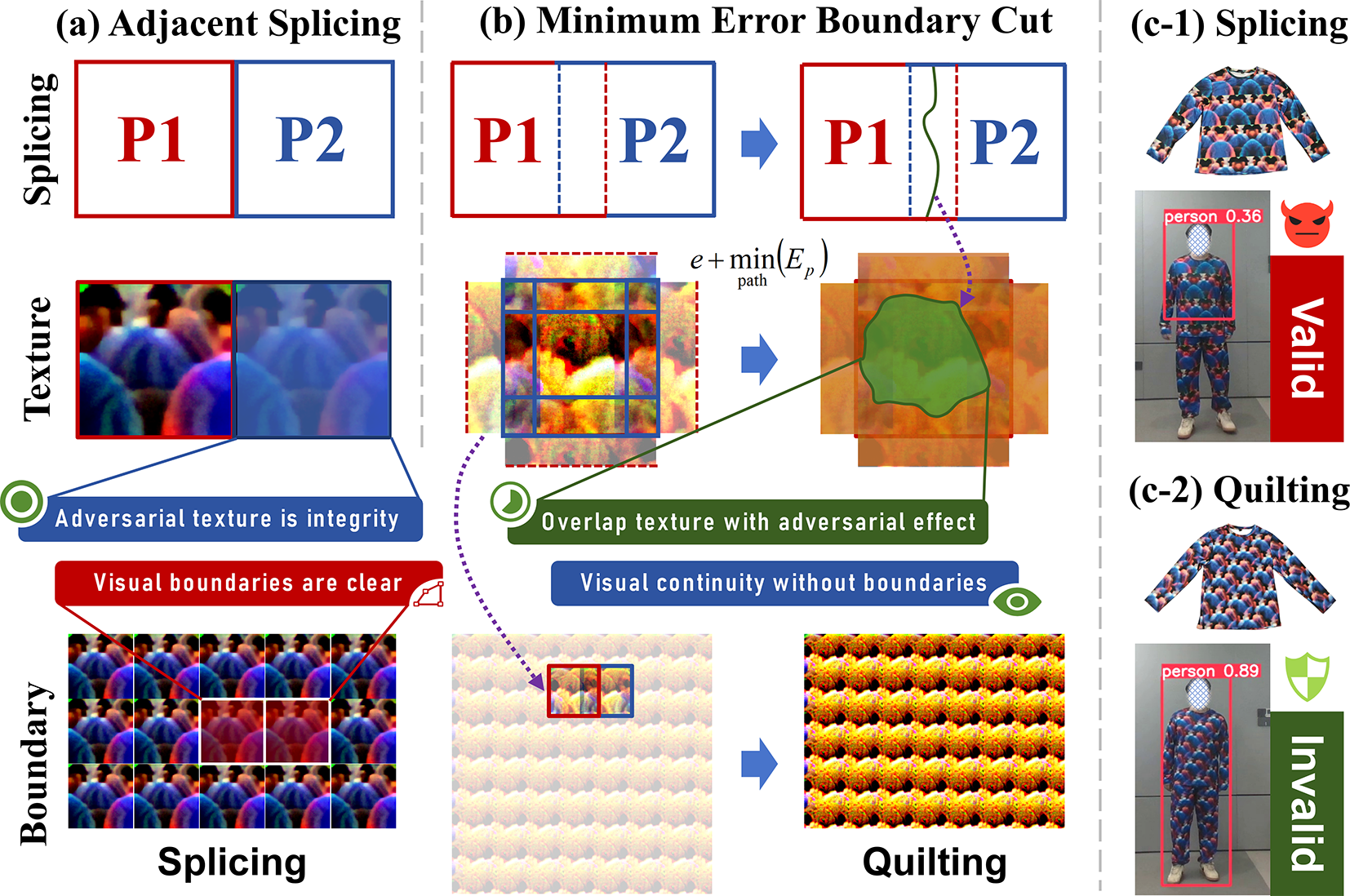}
\caption{Standard splicing exposes visible seams, direct quilting tiled neutralizes adversarial potency, and the proposed Minimum Error Boundary Cut eliminates boundaries while preserving attack effectiveness.}
\label{quilting_alg}
\vspace{-6 mm} 
\end{figure}

Repeating the adversarial pattern across the clothing surface improves ASR by increasing exposure, but introduces two problems: (i) high-contrast seams between tiles can be detected by boundary-aware defenses such as NAPGuard~\cite{Wu_2024_CVPR}, and (ii) sharp seam artifacts inject non-discriminative activations at tile edges, partially undermining global semantic suppression. As shown in Fig.~\ref{quilting_alg}(a), adjacent patches maintain internal texture integrity yet exhibit clear visual boundaries. We therefore seek a quilting strategy that eliminates visible seams while preserving adversarial potency.

The standard approach in computer graphics is the minimum-error boundary cut (Fig.~\ref{quilting_alg}(b)): given two overlapping tiles $P_1$ and $P_2$, dynamic programming finds the seam $\mathcal{S}$ that minimizes the cumulative pixel-wise squared error $e_{i,j}=\|P_1(i,j)-P_2(i,j)\|^2$ across the overlap region $\Omega_{\text{ov}}$. However, applying this technique off-the-shelf to adversarial patches destroys the optimized pixel distributions and neutralizes the attack (Fig.~\ref{quilting_alg}(c)).

A deeper issue is that pixel-level smoothness does not guarantee feature-level smoothness. As established by Gatys~et~al.~\cite{gatys2015texture}, the perceptual identity of a texture resides in deep feature statistics rather than raw pixel values. Moreover, empirical studies on deep networks~\cite{khromov2024lipschitz} show that local Lipschitz constants $L_\ell$ are spatially non-uniform: in high-$L_\ell$ regions, even small pixel differences can be amplified into substantial feature-space jumps. A seam optimized purely in pixel space may therefore introduce feature discontinuities that compromise the adversarial effect. Therefore, we proposed a novel startegy called Feature Smooth Quilting (FSQ) to solve this gap.

\textbf{Feature-sensitivity weighted quilting.}\quad Motivated by recent work that incorporates detector gradient information into adversarial texture optimization~\cite{liang2025gradient,liang2025grac}, we weight the seam-selection error by the feature sensitivity of each texture location. Let $\delta$ denote the current adversarial texture. We define the sensitivity map as the gradient magnitude aggregated over detector layers and a fixed batch of $B$ training frames:
\begin{equation}
\label{eq:sensitivity}
w_{i,j} = \sum_{k=1}^{L}\left\|\frac{\partial\,\mathcal{L}_{\mathrm{det}}}{\partial\,\delta_{i,j}}\right\|_{k},
\end{equation}
where $L$ denotes the number of detection head layers, $k$ indexes each layer, and the gradient magnitude is averaged over a fixed batch of $B$ training frames before aggregation across layers. And we replace the standard quilting error with a feature-weighted variant:
\begin{equation}
\label{eq:weighted_error}
e^{\mathrm{feat}}_{i,j} = \bigl(1 + \mu\,\hat{w}_{i,j}\bigr)\,\|P_1(i,j)-P_2(i,j)\|^2,
\end{equation}
where $\hat{w}_{i,j}$ is the min-max normalized sensitivity and $\mu\ge 0$ controls the weighting strength ($\mu=0$ recovers standard quilting). The dynamic-programming seam then minimizes the cumulative $e^{\mathrm{feat}}_{i,j}$, routing through regions where the detector's feature response is locally insensitive. Under the assumption that gradient magnitude serves as a proxy for local Lipschitz sensitivity~\cite{liang2025gradient}, seams placed in low-gradient regions incur smaller feature perturbations at boundaries, thereby preserving both visual continuity and adversarial potency. To empirically validate this, Fig.~\ref{Fig11_1} shows that AdvSerial with FSQ maintains robust suppression across all postures after quilting, whereas AdvReal without gradient-weighted seam selection loses effectiveness, confirming that routing seams through low-sensitivity regions reduces adversarially critical feature discontinuities in practice. In practice, the sensitivity map is refreshed every $K$ epochs via a single backward pass, incurring negligible overhead.

\textbf{Cross-seam feature continuity bound.}\quad We now formalize why feature-weighted seam selection provides tighter control over cross-seam feature discontinuities than standard pixel-domain quilting. Let $P_{\text{h}}$ denote the stitched result along an optimal seam, $x[P]$ the rendered image, $F_\ell(\cdot)\in\mathbb{R}^{C_\ell\times H_\ell\times W_\ell}$ the feature map at stage $\ell$ of the detector, and $\Omega_{\text{band}}$ a narrow band around the seam with image-plane projection $\Pi(\cdot)$. For paired samples $p^\pm$ on opposite sides of the seam, the local Lipschitz property of the detector gives:
\begin{equation}
\label{eq:lipschitz}
\|\Delta F_\ell(p)\| \le L_\ell(p)\,\|x[P_{\text{h}}](p^+) - x[P_{\text{h}}](p^-)\|,
\end{equation}
where $L_\ell(p)>0$ denotes the spatially varying local Lipschitz constant at stage $\ell$. Recent empirical studies on neural network Lipschitz behavior~\cite{khromov2024lipschitz} have established that $L_\ell(p)$ is highly non-uniform across spatial locations, with high-$L_\ell$ regions corresponding to areas where the network response is most sensitive to local input perturbations. The soft-blending rasterization in PyTorch3D~\cite{liu2019softras,ravi2020pytorch3d} ensures that the rendering operator is locally smooth, yielding:
\begin{equation}
\label{eq:pixelbound}
\|x[P_{\text{h}}](p^+)-x[P_{\text{h}}](p^-)\| \le C_{\text{cam}}\;\max_{(i,j)\in\Omega_{\text{ov}}}\sqrt{e_{i,j}},
\end{equation}
where $C_{\text{cam}}>0$ accounts for projection geometry and shading. Combining Eq.~\ref{eq:lipschitz} and Eq.~\ref{eq:pixelbound} yields the cross-seam feature continuity bound:
\begin{equation}
\label{eq:featurebound}
\|\Delta F_\ell(p)\| \le L_\ell(p)\,C_{\text{cam}}\;\max_{(i,j)\in\Omega_{\text{ov}}}\sqrt{e_{i,j}},\qquad p\in\Pi(\Omega_{\text{band}}).
\end{equation}

\textbf{Bound-tightening property of FSQ.}\quad Let $\mathcal{S}_{\mathrm{std}}=\arg\min_{\mathcal{S}}\sum_{(i,j)\in\mathcal{S}}e_{i,j}$ denote the seam selected by standard pixel-domain quilting, and $\mathcal{S}_{\mathrm{feat}}=\arg\min_{\mathcal{S}}\sum_{(i,j)\in\mathcal{S}}(1+\mu\hat{w}_{i,j})\,e_{i,j}$ the seam selected by FSQ. Define the worst-case cross-seam feature bound along seam $\mathcal{S}$ as:
\begin{equation}
\label{eq:phi_bound}
\Phi_\ell(\mathcal{S}) = C_{\text{cam}}\cdot\max_{(i,j)\in\mathcal{S}}\Bigl[L_\ell(i,j)\cdot\sqrt{e_{i,j}}\Bigr].
\end{equation}
Then under the empirical observation that the gradient magnitude $w_{i,j}$ and the local Lipschitz constant $L_\ell(i,j)$ both quantify the local sensitivity of the detector's response to texture perturbations at position $(i,j)$ and exhibit positive correlation~\cite{khromov2024lipschitz}, the FSQ seam satisfies:
\begin{equation}
\label{eq:tighter_bound}
\Phi_\ell(\mathcal{S}_{\mathrm{feat}}) \le \Phi_\ell(\mathcal{S}_{\mathrm{std}}).
\end{equation}

Intuitively, standard quilting minimizes only the pixel-error factor $\sqrt{e_{i,j}}$ while remaining indifferent to where $L_\ell(i,j)$ is large, leaving the cross-seam feature bound loose in high-$L_\ell$ regions even when the seam is locally pixel-optimal. FSQ applies a multiplicative penalty $(1+\mu\hat{w}_{i,j})$ that grows in regions where the detector's response is most sensitive, steering the seam away from high-$L_\ell$ areas and jointly minimizing both factors that determine the bound on $\|\Delta F_\ell(p)\|$. However, a tighter upper bound alone cannot guarantee small feature discontinuity in practice. The sufficient condition arises from end-to-end joint optimization: $\mathcal{L}_{\mathrm{det}}$ and $\mathcal{L}_{\mathrm{serial}}$ operate directly on the detector's feature responses, and their gradients naturally drive the texture toward cross-seam feature consistency. FSQ constrains the search space to seams with provably tight feature bounds, and the joint optimization then ensures the texture maintains coherent adversarial features across those seams.

The design aligns with broader adversarial theories. FSQ dismantles local shape continuity to achieve semantic suppression~\cite{burgert2026imagenet}, and routes seams through low-sensitivity regions to maintain the low-frequency smoothness that underpins transferability and physical robustness~\cite{dabouei2020smoothfool}. The overall decomposition explains the inadequacy of isolated components. Naive tiling introduces high-frequency boundary artifacts (Fig.~\ref{quilting_alg}(a)). Standard quilting destroys shape-suppressing patterns (Fig.~\ref{quilting_alg}(c-2)). FSQ combined with joint optimization preserves visual continuity and maximizes semantic suppression.


\begin{figure*}[b!]
\centering
\includegraphics[width=15cm]{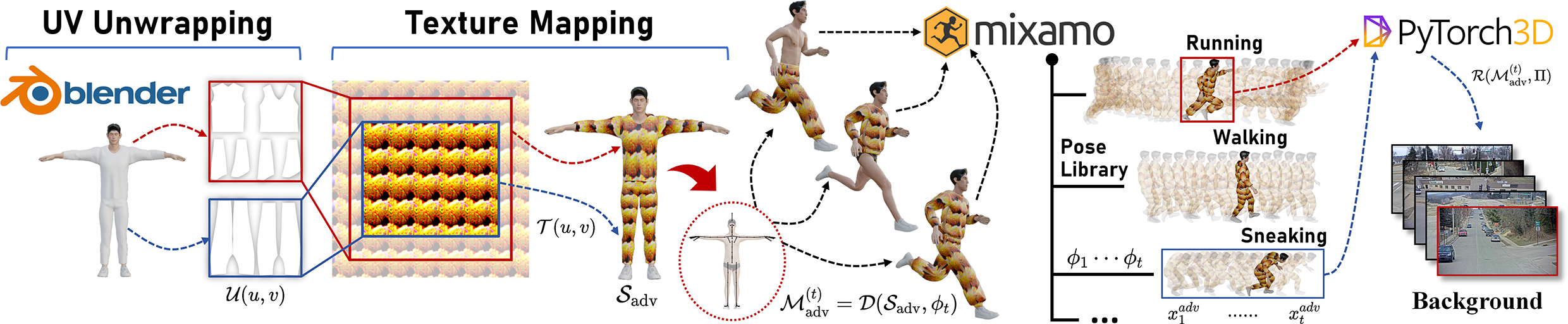}
\caption{The 3D neural rendering pipeline. Illustration of UV unwrapping, adversarial texture mapping onto garment meshes, pose-driven deformation, and differentiable rendering with background compositing.}
\label{pose_gen}
\vspace{-4 mm} 
\end{figure*}

\subsection{Pose Generation}
\label{poseture}

We mitigate this by projecting the quilted texture onto 3D garments with a differentiable renderer, enabling spatially aware adversarial training. As shown in Fig.~\ref{pose_gen}, our pipeline maps the adversarial texture to clothing, animates the dressed mesh with human motion, and renders frames in photorealistic backgrounds. It comprises four stages: UV unwrapping, texture projection, pose-driven deformation, and differentiable rendering with background composition.

\subsubsection{\textbf{UV unwrapping and texture mapping}} 
Let $\delta \in \mathbb{R}^{H \times W \times 3}$ denote the adversarial texture generated through the quilting module. Following~\cite{huang2026advreal}, we first utilize Blender to extract the UV parameterization $\mathcal{U}$ of the 3D garment mesh $\mathcal{M}_{\text{cloth}}$. Subsequently, we employ the differentiable rendering pipeline in PyTorch3D to map the 2D texture $\delta$ onto the 3D surface using $\mathcal{U}$, yielding the textured adversarial surface $\mathcal{S}_{\text{adv}}$. This approach securely binds the 2D pattern to the 3D geometry while maintaining end-to-end differentiability, ensuring the spatial continuity of the adversarial pattern when wrapped around the complex clothing.

\subsubsection{\textbf{Dynamic poses}} 
Given a base clothed human mesh $\mathcal{M}_{\text{hum}}$ (integrated with $\mathcal{S}_{\text{adv}}$) and an animation sequence $\{ \phi_1, \phi_2, \dots, \phi_T \}$ representing skeletal pose parameters (e.g., joint rotations and global body orientation), we apply Linear Blend Skinning (LBS) to articulate the 3D model. For each time step $t$, driven by the specific pose $\phi_t$, the LBS deformation dynamically outputs the adversarially dressed human mesh $\mathcal{M}_{\text{adv}}^{(t)}$. This process introduces critical real-world variations such as limb bending, self-occlusion, and fabric distortion, which serve as essential physical constraints for optimizing temporal robustness.


\subsubsection{\textbf{Human-Background Rendering}}

To generate realistic adversarial frames, the animated 3D human model is projected into the image domain using a differentiable renderer \(\mathcal{R}\). Camera extrinsics are defined by the parameterized pose \(C_t = (R_{\text{cam}, t}, \vec{v}_{\text{cam}, t})\), where \(R_{\text{cam}, t}\) denotes the rotation matrix and \(\vec{v}_{\text{cam}, t}\) represents the translation vector. Adjusting \(C_t\) across time enables simulation of dynamic views and observer motion.

To enhance the realism and variability of the rendered outputs, we generate a randomized lighting configuration $\mathcal{L}$ for each sequence, where the number, type, orientation, position, and intensity of light sources are uniformly sampled from predefined physically plausible ranges.

Once the camera and lighting are specified, the animated mesh $\mathcal{M}_{\text{adv}}^{(t)}$ is projected into the image domain via a differentiable renderer $\mathcal{R}$ by PyTorch3D. The final adversarial frame is synthesized via alpha blending:
\begin{equation}
x_t^{\text{adv}} = M_t \odot \mathcal{R}(\mathcal{M}_{\text{adv}}^{(t)}, \Pi) + (1 - M_t) \odot x_b
\end{equation}
where $\Pi$ is the camera projection operator, $x_b \in \mathbb{R}^{H \times W \times 3}$ a sampled background, and $M_t \in \{0,1\}^{H \times W}$ the silhouette mask from depth testing. 

To simulate motion dynamics, the same adversarial texture \(\delta\) is applied across a temporally coherent pose sequence, producing a video clip $\mathcal{X}^{\text{adv}}$. The sequence models continuous pedestrian motion and supports temporal loss computation. As shown in stages \textcircled{2} and \textcircled{3} of Fig.~\ref{structure}, the rendering process enables robust adversarial training under diverse spatial and temporal conditions.

\subsection{Loss Function}
\label{loss}

\subsubsection{\textbf{Detection Loss $\mathcal{L}_{\text{det}}$}}

Detection loss measures the confidence assigned to the most spatially and semantically valid pedestrian prediction~\cite{xu2020adversarial} and is defined as
\begin{equation}
\mathcal{L}_{\text{det}}(p) = f_\theta(x^{\text{adv}}, b^{*}),  \quad 
b^{*} = \arg\max_{b \in \mathcal{B}} \text{IoU}(b, b_{\text{gt}}),
\end{equation}
where $f_\theta(x^{\text{adv}}, b^*)$ is the confidence score of the adversarial image $x^{\text{adv}}$ at the bounding box $b^*$ that has the highest IoU with the ground truth $b_{\text{gt}}$. Focusing the loss on this high-IoU region yields more stable gradients and stronger degradation of detector performance.


\subsubsection{\textbf{Serial Frame Loss $\mathcal{L}_{serial}$}}
To promote temporally consistent suppression, we design a serial frame loss that penalizes target recovery after successful suppression sequences.

\noindent\textbf{Soft Suppression Score: } 
Detection pipelines typically produce hard binary decisions (detected or not) that block gradient flow. To circumvent this, we define a differentiable success score $\hat{S}_t \in (0,1)$ using the raw objectness confidence $O(f_t, p_t)$ prior to any post-processing:
\begin{equation}
\hat{S}_t = \sigma\left( \beta \cdot (\tau - O(f_t, p_t)) \right),
\label{eq:soft_score}
\end{equation}
where $\tau$ is the detection threshold and $\beta$ controls transition sharpness. $\hat{S}_t \to 1$ implies successful suppression (confidence below $\tau$), and $\hat{S}_t \to 0$ implies detection survival. The function serves as a differentiable proxy for the probability of invisibility and ensures that the optimization process can sense and reduce confidence even when the object approaches the detection decision boundary.

\noindent\textbf{Serial Loss: }
We formulate the loss as a weighted failure rate over $T$ frames:
\begin{equation}
\mathcal{L}_{\text{serial}}
= 1 - \frac{\sum_{t=1}^{T} w_t \hat{S}_t}{\sum_{t=1}^{T} w_t}.
\label{eq:serial-loss}
\end{equation}
The objective acts as a sequence-level reliability metric that focuses optimization on maintaining a high invisibility rate by weighting the importance of each moment in the sequence rather than treating every frame equally.

To favor continuous suppression over intermittent failures, the adaptive weight $w_t$ is defined as:
\begin{equation}
\label{eq:serial-weight}
w_t = \alpha^{T - t} \left( 1 + \lambda \sum_{i=1}^{t-1} \prod_{j=i}^{t-1} \hat{S}_j \right).
\end{equation}
Here $\alpha$ controls temporal decay, while the inner product $\prod_{j=i}^{t-1}\hat{S}_j$ tracks unbroken suppression streaks. It stays close to $1$ only when every frame from $i$ to $t{-}1$ is successfully suppressed, and collapses toward $0$ once any intermediate frame fails. Consequently, $w_t$ grows monotonically with streak length, so a re-detection at frame $t$ after $k$ consecutive successes incurs a penalty roughly proportional to $k$. The strategy creates an asymmetric cost structure in which breaking a long streak is far more expensive than an early isolated failure.

\begin{figure*}[t!]
\centering
\includegraphics[width=16cm]{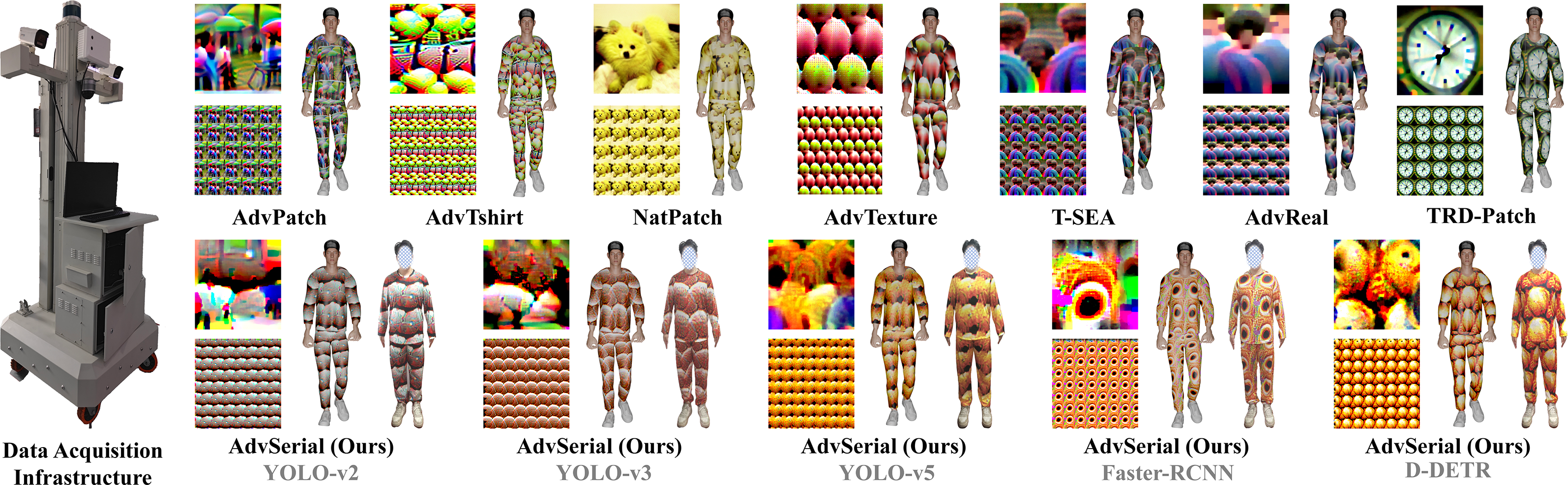}
\caption{Data acquisition infrastructure and adversarial patches. (1) Top row: Patches generated by baseline methods. (2) Bottom row: AdvSerial patches optimized for different white-box detectors.}
\label{allpatch}
\vspace{-4 mm} 
\end{figure*}

From an optimization perspective, this asymmetric penalty acts as an implicit temporal-smoothness regularizer. Image-based adversarial attacks inherently lack temporal continuity constraints~\cite{pony2021over, kim2023breaking}, and unconstrained optimization tends to converge to sharp, isolated minima in the loss landscape. Minor physical perturbations (cloth deformation, micro-movements) can then push the input out of such minima, causing abrupt confidence spikes known as "adversarial flickering"~\cite{pony2021over}. Because the streak-dependent $w_t$ in Eq.~\eqref{eq:serial-weight} effectively penalizes the temporal derivative of the objectness score, it steers the optimizer away from fragile discontinuous solutions and toward flat, temporally shared minima that remain robust across consecutive frames. The convergence behavior and temporal smoothness properties of SFL are empirically validated through ablation studies and hyperparameter sensitivity analysis in Sec.~\ref{ablation}.

\subsubsection{\textbf{Total Variation Loss and Total Loss}}

To suppress unnatural visual discontinuities, we apply the standard Total Variation (TV) loss $\mathcal{L}_{\text{tv}}$, which ensures spatial coherence across the quilted adversarial patch and improves visual quality under physical projection. The overall training objective combines the spatial detection, temporal serial, and regularization terms into a single loss function: $\mathcal{L}_{\text{total}} = \lambda_{\text{det}} \mathcal{L}_{\text{det}} + \lambda_{\text{ser}} \mathcal{L}_{\text{serial}} + \lambda_{\text{tv}} \mathcal{L}_{\text{tv}}$. The balancing weights are empirically set to $\lambda_{\text{det}} = 0.5$, $\lambda_{\text{ser}} = 0.5$, and $\lambda_{\text{tv}} = 0.1$ to effectively control the strength of temporal consistency and spatial smoothness.

\section{Experiments}
\label{experiments}

\subsection{Experiments Setup}

\begin{table*}[t]
\centering
\caption{Specifications of victim detectors. Summary of the architecture type (Stage, Anchor-based/free), specific variant, backbone, and GFLOPs for the models used in evaluation.}
\label{detector}
\resizebox{12cm}{!}{
\renewcommand{\arraystretch}{1.1} 

\begin{tabular}{cccccccc} 
\hline
\textbf{Detector} & \textbf{Stage} & \textbf{Anchor} & \textbf{Variant} & \textbf{Backbone} & \textbf{GFLOPs} & \textbf{Years} & \textbf{Cites} \\ \hline
\textbf{YOLO-v2} & \multirow{6}{*}{One-Stage} & \multirow{3}{*}{Anchor-based} & Standard & Darknet-19 & 5.58 & 2017 & \cite{redmon2017yolo9000} \\
\textbf{YOLO-v3} &  &  & Standard & Darknet-53 & 139.89 & 2018 & \cite{redmon2018yolov3incrementalimprovement} \\
\textbf{YOLO-v5n} &  &  & Small & CSPDarknet & 7.70 & 2020 & \cite{yolov5} \\ \cline{3-8} 
\textbf{YOLO-v8n} &  & \multirow{3}{*}{Anchor-free} & Nano & C2f-Darknet & 8.70 & 2023 & \cite{yolov8_ultralytics} \\
\textbf{YOLO-v10n} &  &  & Nano & CSPNet & 6.50 & 2024 & \cite{wang2024yolov10} \\
\textbf{YOLO-v12n} &  &  & Nano & Area-Attention & 6.50 & 2025 & \cite{tian2025yolov12} \\ \cline{2-8} 
\textbf{Faster-RCNN} & \multirow{2}{*}{Two-Stage} & Anchor-based & Standard & ResNet-50-FPN & 180.00 & 2016 & \cite{ren2016fasterrcnnrealtimeobject} \\ \cline{3-8} 
\textbf{D-DETR} &  & Anchor-free & Standard & ResNet-50 & 78.00 & 2020 & \cite{ddetr} \\ \hline
\end{tabular}
}
\vspace{-4 mm} 
\end{table*}

\begin{table*}[b]
\renewcommand{\arraystretch}{1.15} 
\centering
\caption{Attack Success Rate (ASR) comparison. Evaluation of baseline and proposed methods on YOLO-v2 and Faster-RCNN across different IoU thresholds. All adversarial patches were generated using YOLO-v2 as the surrogate model, as it is the only common target supported by the official assets of all baselines.}
\label{comparison_all}
\resizebox{14cm}{!}{
\begin{tabular}{ccccccccccc}
\hline
& \multicolumn{5}{c}{\textbf{YOLO-v2 (ASR $\uparrow$)}} & \multicolumn{5}{c}{\textbf{Faster-RCNN (ASR $\uparrow$)}} \\ 
\cline{2-11}
\multirow{-2}{*}{} & IoU=0.1 & IoU=0.3 & IoU=0.5 & IoU=0.7 & \multicolumn{1}{c|}{IoU=0.9} & IoU=0.1 & IoU=0.3 & IoU=0.5 & IoU=0.7 & IoU=0.9 \\ \hline

\textbf{White}         & 14.29\% & 14.29\% & 16.95\% & 42.29\% & \multicolumn{1}{c|}{98.86\%} & 5.14\%  & 5.14\%  & 5.52\%  & 12.95\% & 72.29\% \\
\textbf{Gray}          & 12.76\% & 12.76\% & 13.33\% & 31.43\% & \multicolumn{1}{c|}{97.52\%} & 4.76\%  & 4.76\%  & 5.90\%  & 10.29\% & 69.90\% \\
\textbf{Random}        & 14.29\% & 14.29\% & 14.29\% & 28.00\% & \multicolumn{1}{c|}{96.76\%} & 4.95\%  & 4.95\%  & 6.29\%  & 11.43\% & 71.05\% \\ \hline

\textbf{AdvPatch}      & 52.38\% & 52.38\% & 52.57\% & 58.10\% & \multicolumn{1}{c|}{96.76\%} & 20.95\% & 21.33\% & 25.71\% & 44.76\% & 88.95\% \\
\textbf{AdvTshirt}     & 49.90\% & 49.90\% & 50.48\% & 58.10\% & \multicolumn{1}{c|}{97.71\%} & 19.81\% & 20.00\% & 24.57\% & 43.05\% & 88.95\% \\
\textbf{NatPatch}      & \cellcolor{lightyellow}68.95\% & \cellcolor{lightyellow}68.95\% & \cellcolor{lightyellow}69.14\% & \cellcolor{lightyellow}74.48\% & \multicolumn{1}{c|}{\cellcolor{lightyellow}99.43\%} & 22.86\% & 22.48\% & 27.43\% & 48.95\% & 91.62\% \\
\textbf{AdvTexture}    & 40.00\% & 40.00\% & 40.00\% & 47.62\% & \multicolumn{1}{c|}{95.43\%} & 15.62\% & 16.00\% & 20.38\% & 39.24\% & 87.81\% \\
\textbf{TSEA}       & 45.52\% & 45.71\% & 46.29\% & 52.57\% & \multicolumn{1}{c|}{96.76\%} & 18.48\% & 18.67\% & 22.86\% & 40.19\% & 89.00\% \\
\textbf{TRDPatch}     & 64.57\% & 64.57\% & 64.76\% & 69.52\% & \multicolumn{1}{c|}{99.05\%} & \cellcolor{lightyellow}24.57\% & 24.95\% & 28.38\% & 50.67\% & 92.19\% \\
\textbf{AdvReal}    & 53.90\% & 53.90\% & 54.10\% & 59.24\% & \multicolumn{1}{c|}{96.57\%} & 22.10\% & \cellcolor{lightyellow}27.05\% & \cellcolor{lightyellow}30.29\% & \cellcolor{lightyellow}53.52\% & \cellcolor{lightgreen}\textbf{93.14\%} \\ \hline

\textbf{AdvSerial} & \cellcolor{lightgreen}\textbf{89.71\%} & \cellcolor{lightgreen}\textbf{89.71\%} & \cellcolor{lightgreen}\textbf{89.71\%} & \cellcolor{lightgreen}\textbf{91.81\%} & \multicolumn{1}{c|}{\cellcolor{lightgreen}\textbf{99.43\%}} & \cellcolor{lightgreen}\textbf{76.62\%} & \cellcolor{lightgreen}\textbf{76.62\%} & \cellcolor{lightgreen}\textbf{79.24\%} & \cellcolor{lightgreen}\textbf{80.19\%} & \cellcolor{lightyellow}92.76\% \\ \hline
\end{tabular}
}
\end{table*}

\textbf{Training Details: }
To ensure rigorous reproducibility, we detail our experimental setup and hyper-parameters. All experiments are conducted on an NVIDIA A800 GPU (80GB). The adversarial patterns are optimized for 1500 epochs using the Adam optimizer with an initial learning rate of 0.001. The input image resolution is set to $416 \times 416$. For the quilting module, we utilize an initial patch block size of $80 \times 80$, an overlap margin of 25, a scaling factor of 5, and an error tolerance of 0.3. The generated texture is mapped to cover approximately $85\%$ of the human body and rendered under randomly sampled viewpoints (angles and positions). Crucially, to compute the serial loss ($\mathcal{L}_{serial}$) and maintain temporal consistency, we extract 8 continuous frames per sequence (i.e., Serial Frame Length, SFL = 8). Consequently, our overall spatial batch size of 64 is formulated as a 3D batch comprising 8 independent temporal sequences ($8 \text{ sequences} \times 8 \text{ frames}$). The final objective weights are empirically set to $\lambda_{det}=0.5$, $\lambda_{ser}=0.5$, and $\lambda_{tv}=0.1$.

\textbf{Adversarial Attack Baselines.} 
We compared our method \textit{AdvSerial} with both classic and state-of-the-art (SOTA) adversarial methods. The classic methods include \textit{AdvPatch}~\cite{thys2019fooling}, \textit{AdvTshirt}~\cite{xu2020adversarial}, and \textit{NatPatch}~\cite{hu2021naturalistic}, while the SOTA methods comprise \textit{AdvTexture}~\cite{hu2022adversarial}, \textit{T-SEA}~\cite{huang2023t}, \textit{AdvReal}~\cite{huang2026advreal}, and \textit{TRD-Patch}~\cite{wang2025transferable}. In addition, we assess \textit{White}, \textit{Gray}, and \textit{Random Noise} patches as control groups. To avoid implementation bias and ensure optimal baseline performance, we directly utilize the official pre-trained adversarial patches released by the original authors. Consequently, the evaluation is conducted on YOLO-v2, as it represents the only common target model supported by the official assets of all comparison methods. The areas of all patches on both rendered and real garments are kept consistent across digital and physical experiments. The remaining patches and their corresponding garments are shown in Fig.~\ref{allpatch}.

\textbf{Datasets.}
The adversarial texture is trained using two data sources. In the 2D branch, we use the \href{https://universe.roboflow.com/pascal-to-yolo-8yygq/inria-person-detection-dataset}{INRIA person dataset} (462 training images). In the 3D branch, we randomly select $1{,}538$ background images from the \href{https://www.aicitychallenge.org/}{AI City 2022 dataset} for rendering training samples. The AI City dataset contains roadside infrastructure footage from multiple urban perspectives, which aligns with our target scenario of infrastructure-based high-angle monitoring. For digital evaluation, we use a separate set of $462$ background images from AI City. For each adversarial patch, $8$ rendered samples are generated per background image, yielding a total of $3{,}696$ test images per method. For the defense evaluation against Sparse4D-v3, we follow its standard benchmark protocol on the \href{https://www.nuscenes.org/}{nuScenes-v1.0mini} dataset. Physical-world evaluation is conducted using real printed garments and a Hikvision surveillance camera (1/2.7'' Progressive Scan CMOS sensor, $1920\times1080$ resolution, 4\,mm focal length, $87.6^\circ$ horizontal field of view, F1.6 aperture, 25\,fps) mounted on a height-adjustable rig. Data are collected at mounting heights of $3\,$m in the indoor setting and $5\,$m in the outdoor setting. Combined with subject distances ranging from $5$ to $30\,$m, the outdoor configuration covers pitch angles from approximately $9.5^\circ$ to $45^\circ$, representative of the elevated viewpoints encountered in infrastructure-mounted surveillance deployments. The acquisition setup is illustrated in Fig.~\ref{allpatch} and comprehensive experimental details are provided in Sec.~\ref{sec:physical}.

\textbf{Victim Detectors.}
We evaluate adversarial patches in physical world across mainstream object detectors encompassing both single-stage and two-stage paradigms, as well as anchor-based and anchor-free designs (Tab.~\ref{detector}). Specifically, the benchmark includes classical YOLO series (YOLO-v2, YOLO-v3, and YOLO-v5n), SOTA anchor-free variants (YOLO-v8n, YOLO-v10n, YOLO-v12n), and representative architectures such as Faster R-CNN and D-DETR. All detectors are pretrained on the MS COCO dataset~\cite{lin2014microsoft}. Generally, models with higher GFLOPs exhibit greater robustness and generalization capability against adversarial perturbations.

\textbf{Evaluation Metrics.}
We employ standard detection metrics, including ASR, Recall, Precision, and F1-score. Lower values in Recall, Precision, and F1-score indicate more severe degradation of the victim detector under attack~\cite{huang2026advreal}.

The subsequent experiments are organized to validate each design component and the system-level impact of AdvSerial against infrastructure-based surveillance pipelines. Sec.~\ref{sec:comparison} establishes the fundamental semantic suppression capability against diverse SOTA detectors. Sec.~\ref{sec:transfer} validates cross-model generalizability and directly connects sustained frame-level suppression to system-level disruption of downstream tracking pipelines. Sec.~\ref{sec:ablation} quantifies the contribution of each design component and confirms the stability of the serial frame loss optimization. Sec.~\ref{sec:robustness} validates the high-angle and all-weather robustness that infrastructure-mounted cameras demand, including pitch-angle simulation and defense evaluation against both boundary-aware patch detectors and 3D-temporal fusion models. Sec.~\ref{sec:physical} validates the end-to-end framework under realistic infrastructure deployment conditions across varying distances, azimuths, and human postures. Sec.~\ref{sec:explainability} provides mechanistic insight into how the adversarial texture disrupts person-specific feature representations within the detector.


\subsection{Comparisons to SOTA Adversarial Attack Methods}
\label{sec:comparison}

To comprehensively evaluate AdvSerial, we compare it with several SOTA adversarial patches under two representative detector architectures.

Unoptimized patterns (White, Gray, Random) in Tab.~\ref{comparison_all} confirm that attack success stems from adversarial optimization, as both detectors maintain ASR below $15\%$ at IoU$\le 0.5$. Among optimized methods, NatPatch and TRDPatch lead on YOLO-v2 with $69.14\%$ and $64.76\%$ ASR at IoU$=$0.5, while AdvReal achieves the highest Faster-RCNN performance ($30.29\%$ at the same threshold).

AdvSerial consistently achieves the highest ASR across nearly all settings. On YOLO-v2, its ASR stays at $89.71\%$ from IoU$=$0.1 to $0.5$ and rises only to $91.81\%$ at IoU$=$0.7. On Faster-RCNN it remains within $76.62\%$--$79.24\%$ over the same range. This stability across IoU thresholds indicates that AdvSerial primarily suppresses person-specific semantic activations rather than disturbing bounding-box localization, making its effectiveness largely independent of the localization criterion.


\subsection{Cross-Model Transferability}
\label{sec:transfer}

\begin{table*}[t!]
\centering
\caption{Cross-model transferability results. Quantitative metrics (Recall, Precision, F1-score) on YOLO-v8n, v10n, and v12n using patches trained on YOLO-v2 (white-box).}
\resizebox{14cm}{!}{
\setlength{\tabcolsep}{4pt}
\renewcommand{\arraystretch}{1.15}
\begin{tabular}{c|ccccccccc}
\hline
\multirow{2}{*}{\textbf{Method}} & \multicolumn{3}{c}{\textbf{YOLO-v8n}}                               & \multicolumn{3}{c}{\textbf{YOLO-v10n}}                               & \multicolumn{3}{c}{\textbf{YOLO-v12n}}                     \\ \cline{2-10} 
                                 & \textbf{Recall $\downarrow$} & \textbf{Precision $\downarrow$} & \textbf{F1-score $\downarrow$}       & \textbf{Recall $\downarrow$} & \textbf{Precision $\downarrow$} & \textbf{F1-score $\downarrow$}       & \textbf{Recall $\downarrow$} & \textbf{Precision $\downarrow$} & \textbf{F1-score $\downarrow$} \\ \cline{1-10} 
\textbf{AdvPatch}                & 92.00\%         & 96.79\%             & \multicolumn{1}{c|}{94.34\%} & 96.65\%         & 96.83\%             & \multicolumn{1}{c|}{96.74\%} & 95.43\%         & 97.66\%             & 96.53\%            \\
\textbf{AdvTshirt}               & 95.62\%         & 97.10\%             & \multicolumn{1}{c|}{96.35\%} & 92.83\%         & 96.48\%             & \multicolumn{1}{c|}{94.62\%} & 97.14\%         & 97.70\%             & 97.42\%            \\
\textbf{NatPatch}                & 98.29\%         & 97.18\%             & \multicolumn{1}{c|}{97.73\%} & 96.97\%         & 96.79\%             & \multicolumn{1}{c|}{96.88\%} & 98.29\%         & 97.73\%             & 98.01\%            \\
\textbf{AdvTexture}              & 96.95\%         & 97.14\%             & \multicolumn{1}{c|}{97.04\%} & 94.64\%         & 96.63\%             & \multicolumn{1}{c|}{95.62\%} & 97.14\%         & 97.70\%             & 97.42\%            \\
\textbf{TSEA}                    & 95.43\%         & 97.09\%             & \multicolumn{1}{c|}{96.25\%} & 97.62\%         & 96.39\%             & \multicolumn{1}{c|}{97.00\%} & 94.67\%         & \cellcolor{lightyellow}97.64\%             & 96.13\%            \\
\textbf{TRDPatch}                & 94.48\%         & 96.87\%             & \multicolumn{1}{c|}{95.66\%} & 97.73\%         & 96.58\%             & \multicolumn{1}{c|}{97.15\%} & 95.43\%         & 97.66\%             & 96.53\%            \\
\textbf{AdvReal}                 & \cellcolor{lightyellow}82.86\%         & \cellcolor{lightyellow}94.36\%             & \multicolumn{1}{c|}{\cellcolor{lightyellow}88.24\%} & \cellcolor{lightyellow}85.88\%         & \cellcolor{lightgreen}{\textbf{95.43\%}}             & \multicolumn{1}{c|}{\cellcolor{lightyellow}90.40\%} & \cellcolor{lightyellow}92.57\%         & \cellcolor{lightgreen}\textbf{97.59\%}             & \cellcolor{lightyellow}95.01\%            \\
\textbf{AdvSerial}               & \cellcolor{lightgreen}\textbf{63.81\%}         & \cellcolor{lightgreen}\textbf{91.78\%}             & \multicolumn{1}{c|}{\cellcolor{lightgreen}\textbf{75.28\%}} & \cellcolor{lightgreen}\textbf{67.05\%}         & \cellcolor{lightyellow}95.93\%             & \multicolumn{1}{c|}{\cellcolor{lightgreen}\textbf{78.92\%}} & \cellcolor{lightgreen}\textbf{92.38\%}         & \cellcolor{lightgreen}\textbf{97.59\%}             & \cellcolor{lightgreen}\textbf{94.91\%}            \\ \hline
\end{tabular}
}
\label{comp_sota}
\vspace{-4 mm} 
\end{table*}

\subsubsection{Cross-Model Attack Performance}
\begin{figure}[t!]
\centering
\includegraphics[width=6 cm]{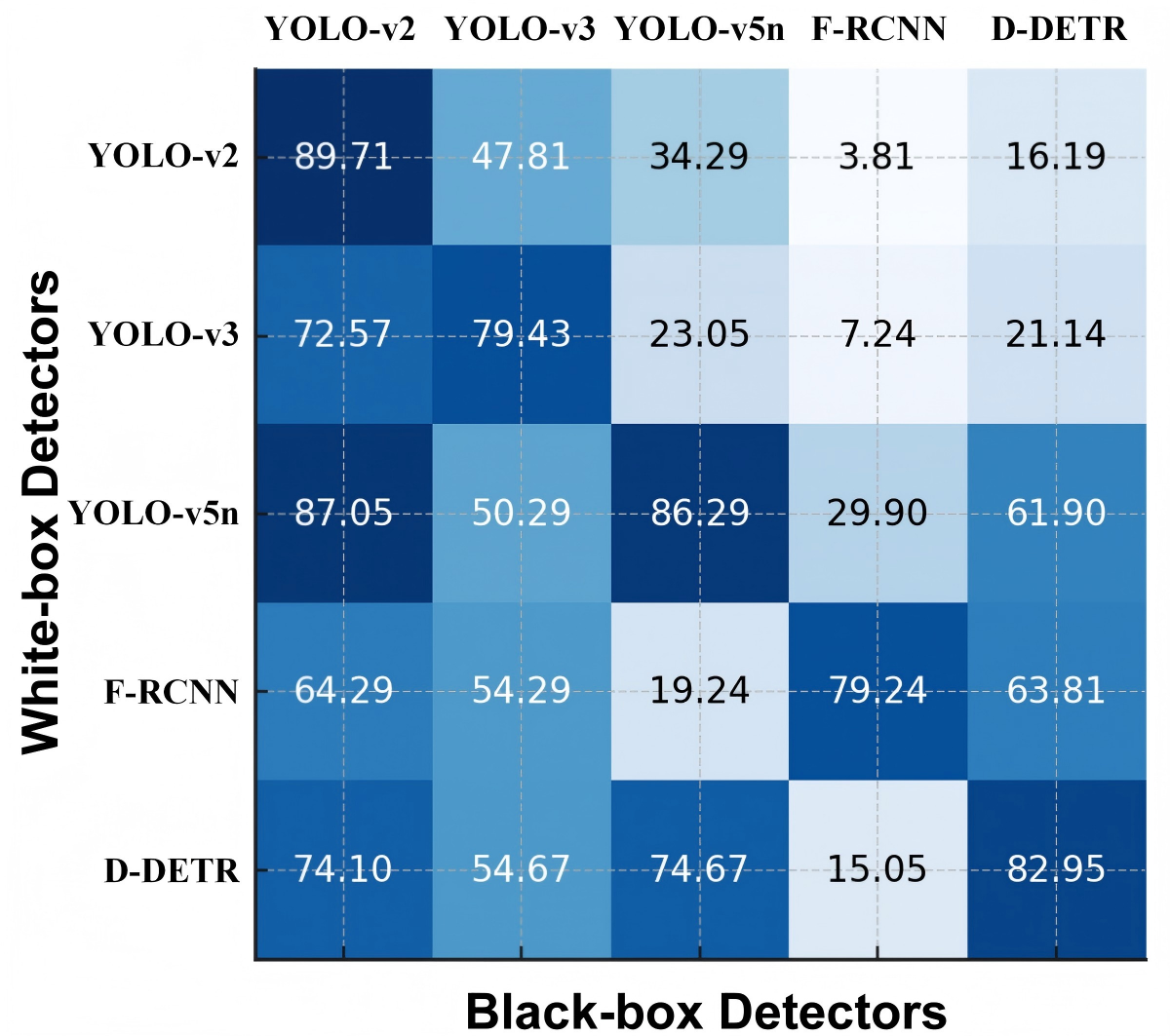}
\caption{Cross-detector transferability matrix. The heatmap displays the ASR when transferring patches trained on white-box models (rows) to black-box models (columns)}
\label{transferability}
\vspace{-4 mm} 
\end{figure}

We first analyze cross-model transferability in black-box settings. Specifically, we evaluate generalization by training a patch on the white-box model and testing it on black-box targets comprising YOLO-v2, YOLO-v3, and YOLO-v5 as one-stage detectors, Faster R-CNN as a two-stage detector, and D-DETR as a transformer-based detector. The asymmetric transfer matrix in Fig.~\ref{transferability} yields three findings. (1) Intra-family transfer is strongest, e.g., the YOLO-v5 patch attains $87.05\%$ ASR on YOLO-v2. (2) Cross-paradigm transfer declines sharply, with YOLO$\to$Faster R-CNN ASR dropping to $3.81\%$--$29.90\%$ due to architectural mismatches. (3) Transfer is markedly asymmetric: D-DETR$\to$YOLO reaches $74.10\%$--$74.67\%$, whereas YOLO$\to$D-DETR stays at $16.19\%$--$61.90\%$.

These patterns indicate that patches primarily exploit family-specific inductive biases, while mismatches in detection heads and training objectives drive the cross-paradigm decline and directional asymmetry.


\subsubsection{Transferability to SOTA Victim Detectors}

To evaluate cross-model generalization, we conduct transferability experiments on three SOTA detectors using YOLO-v2 as the white-box model. Results are summarized in Tab.~\ref{comp_sota}, where lower Recall, Precision, and F1-score indicate stronger transferability.

Classical patches such as AdvPatch, AdvTshirt, and NatPatch show limited generalization, maintaining recall and precision above $95\%$ across all targets. TSEA and TRDPatch achieve slightly better cross-model degradation. In contrast, AdvReal and AdvSerial demonstrate markedly higher transferability. AdvReal reduces F1 to $88.24\%$, $90.40\%$, and $95.01\%$ on YOLO-v8n, v10n, and v12n, respectively, while AdvSerial further decreases F1 to $75.28\%$ on YOLO-v8n and $78.92\%$ on YOLO-v10n. Notably, AdvSerial achieves the lowest overall recall ($63.81\%$ on YOLO-v8n and $67.05\%$ on YOLO-v10n), with its advantage further amplified on the NMS-free YOLO-v10n, as Eq.~\ref{eq:soft_score} and Eq.~\ref{eq:serial-loss} suppresses confidence prior to post-processing where no redundant candidate re-scoring exists to compensate.


\subsubsection{Cross-Task Transferability to Detection and Tracking}

\begin{table*}[t!]
\centering
\caption{Impact on downstream object tracking. Evaluation of tracking stability metrics (ASR-T, ID Switches, Frame Loss Times) on four representative trackers.}
\resizebox{14cm}{!}{
\setlength{\tabcolsep}{4pt}
\renewcommand{\arraystretch}{1}
\begin{tabular}{@{}ccccccccccccc@{}}
\toprule
\multirow{2}{*}{\textbf{Model}} & \multicolumn{3}{c}{\textbf{ASR-T $\uparrow$}} & \multicolumn{3}{c}{\textbf{mIDSW $\uparrow$}} & \multicolumn{3}{c}{\textbf{MFLT $\uparrow$}} & \multicolumn{3}{c}{\textbf{Fragmentation}} \\ \cmidrule(l){2-13} 
                       & \textbf{Clean}         & \textbf{Ours}        & \textbf{AdvReal}        & \textbf{Clean}       & \textbf{Ours}       & \textbf{AdvReal}      & \textbf{Clean}       & \textbf{Ours}       & \textbf{AdvReal}      & \textbf{Clean}   & \textbf{Ours}   & \textbf{AdvReal}   \\ \midrule
\textbf{Deepsort}               & 7.69\%          & 56.72\%            & 28.62\%          & 1.00        & 1.80            & 1.70         & 10       & 74           & 39        & 3.30    & 5.30        & 22.40     \\
\textbf{Bytetrack}              & 8.46\%          & 89.46\%            & 61.47\%          & 1.00        & 2.10            & 1.30         & 8        & 47           & 17        & 3.70    & 9.70        & 13.40     \\
\textbf{Qdtrack}                & 14.62\%         & 29.23\%            & 36.92\%          & 1.00        & 1.00            & 1.00         & 13       & 13           & 12        & 3.10    & 7.90        & 12.70     \\
\textbf{Siamrpn}++              & 4.62\%          & 6.15\%             & 4.62\%           & -           & -               & -            & 6        & 8            & 6         & -       & -           & -         \\ \bottomrule
\end{tabular}
\label{track}
}
\end{table*}

\begin{figure*}[t]
\centering
\includegraphics[width=14 cm]{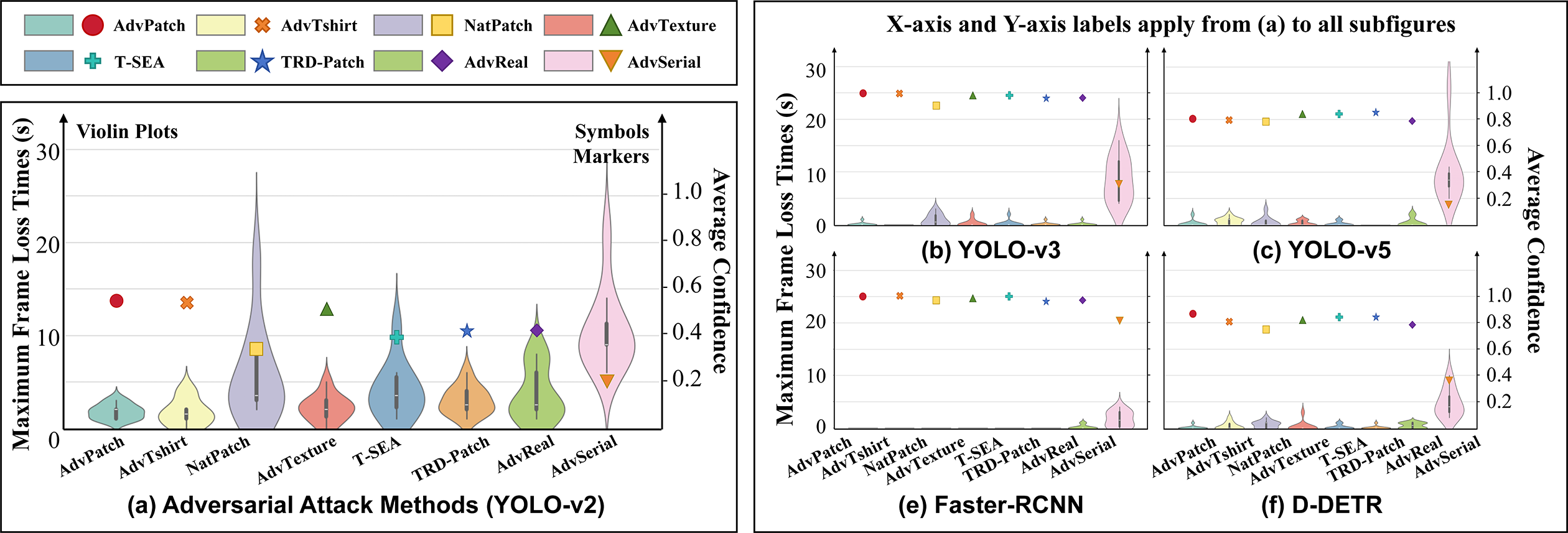}
\caption{Analysis of continuous attack capability. Violin plots displaying the distribution of Maximum Frame Loss Times (duration of continuous failure) and average confidence across simulated video sequences.}
\label{serial}
\vspace{-4 mm} 
\end{figure*}

\textbf{Frame-Based Serial Detection. }
To evaluate robustness in realistic surveillance, we constructed a simulated traffic intersection with a camera height of $4.5\text{m}$. Fig.~\ref{serial} visualizes the attack continuity via \textit{Maximum Frame Loss Times} and average confidence. AdvSerial demonstrates superior temporal transferability on YOLO-v2, sustaining a median consecutive frame loss of $\sim11$ frames (max $\sim29$) and suppressing mean confidence to $0.22$. In contrast, baselines like AdvReal produce only transient flickering with significantly higher confidence ($\approx0.42$). This dominance generalizes across all evaluated architectures (Fig.~\ref{serial} b–e), confirming that AdvSerial creates sustained, temporally consistent detection voids . These long-duration failures are critical for defeating video-based tracking logic, as they effectively sever the temporal association links required by algorithms like DeepSort.

\textbf{Temporal Serial Tracking.}
We evaluate the impact of AdvSerial on downstream tracking tasks using four representative paradigms: DeepSort~\cite{wojke2017simple} (tracking with detection), ByteTrack~\cite{zhang2022bytetrack} (adaptive thresholds), QdTrack~\cite{pang2021quasi} (temporal fusion), and SiamRPN++~\cite{li2019siamrpn++} (siamese tracker). Performance is assessed via Attack Success Rate Tracking (ASR-T), mean identity switches (mIDSW), maximum frame loss times (MFLT), and fragmentation to measure both tracking continuity and stability.

Tab.~\ref{track} reveals a clear pattern consistent with the architectural design of each tracker. AdvSerial is highly effective against detection-dependent pipelines. For DeepSort, the surge in MFLT ($10$ to $74$) confirms effective severance of detection inputs. Against ByteTrack, AdvSerial achieves $89.46\%$ mTSR, outperforming the misleading-oriented AdvReal. Despite ByteTrack's adaptive thresholds, AdvSerial's deep suppression drives objectness scores below the minimum recovery threshold, severing temporal association links. Efficacy decreases against QdTrack ($29.23\%$ mTSR), whose quasi-dense similarity learning leverages temporal priors to partially compensate for frame-level suppression. The detector-free SiamRPN++ remains largely unaffected, as expected, since it operates on template matching without invoking the detection backbone. These results demonstrate that AdvSerial precisely targets the detection stage, which constitutes the foundational layer of modern infrastructure surveillance pipelines. In operational deployments, pedestrian monitoring systems universally rely on detection-based architectures as the primary perception module, with tracking serving as a downstream consumer of detection outputs. The ability to sustain detection failures across consecutive frames therefore directly translates to system-level disruption in real-world infrastructure surveillance scenarios.


\subsection{Ablation Study}
\label{sec:ablation}

To ensure statistical rigor, all ablation configurations are evaluated over 3 independent training runs on YOLO-v5n, with both the mean ASR and standard deviation reported. Tab.~\ref{ablation} ablates three components: Texture Quilting (TQ), Pose Generation (PG), and Serial Frame Loss (SFL). Starting from a baseline ASR of $82.75\%_{\pm 0.58\%}$, introducing any single module yields consistent gains. TQ improves spatial coherence by removing seam artifacts and raises ASR to $84.62\%_{\pm 0.45\%}$ ($+1.87\%$). PG increases view diversity during training and reaches $84.25\%_{\pm 0.52\%}$ ($+1.50\%$). SFL delivers the largest single-module boost, $85.41\%_{\pm 0.41\%}$ ($+2.66\%$), indicating that temporal regularization enhances frame-to-frame attack stability.

Pairwise combinations amplify the effects. TQ+SFL attains $87.55\%_{\pm 0.28\%}$ ($+4.80\%$), TQ+PG reaches $86.35\%_{\pm 0.36\%}$ ($+3.60\%$), and PG+SFL yields $86.22\%_{\pm 0.39\%}$ ($+3.47\%$), showing strong complementarity. Integrating all three modules achieves the best result, $89.85\%_{\pm 0.15\%}$ ($+7.10\%$ over baseline), confirming that spatial quilting, pose diversity, and serial-frame temporal regularization act synergistically to improve the transferability and robustness of the adversarial textures. Furthermore, the decreasing standard deviation (from $\pm 0.58\%$ down to $\pm 0.15\%$) inherently demonstrates that our full pipeline significantly enhances the stability of the adversarial optimization process.

\begin{table}[t!]
\centering
\caption{Ablation study results on YOLOv5n. TQ: Texture Quilting; PG: Pose Generation; SFL: Serial Frame Loss. ASR is reported as mean\,$\pm$\,std over 3 independent runs.}
\label{ablation}
\resizebox{5cm}{!}{%
\setlength{\tabcolsep}{5pt}
\renewcommand{\arraystretch}{1.1}
\begin{tabular}{c ccc r}
\toprule
\textbf{Variant} & \textbf{TQ} & \textbf{PG} & \textbf{SFL} & \textbf{ASR (\%) $\uparrow$} \\
\midrule
\#0 & & & & $82.75 \pm 0.58$ \\
\midrule
\#1 & \checkmark & & & $84.62 \pm 0.45$ \\
\#2 & & \checkmark & & $84.25 \pm 0.52$ \\
\#3 & & & \checkmark & $85.41 \pm 0.41$ \\
\midrule
\#4 & \checkmark & \checkmark & & $86.35 \pm 0.36$ \\
\#5 & & \checkmark & \checkmark & $86.22 \pm 0.39$ \\
\#6 & \checkmark & & \checkmark & $87.55 \pm 0.28$ \\
\midrule
\#7 & \checkmark & \checkmark & \checkmark & $\mathbf{89.85 \pm 0.15}$ \\
\bottomrule
\end{tabular}%
}
\vspace{-4mm}
\end{table}

\subsection{Robustness Study}
\label{sec:robustness}

\begin{table}[t!]
\renewcommand{\arraystretch}{1}
\centering
\caption{Robustness evaluation on PADetBench. Comparison of ASR and AP under complex environmental conditions, including varying weather and distances.}
\label{weather}
\resizebox{8.6cm}{!}{
\begin{tabular}{@{}ccccccccccc@{}}
\toprule
 & \multicolumn{2}{c}{\textbf{Sunny Day}} & \multicolumn{2}{c}{\textbf{Clear Night}} & \multicolumn{2}{c}{\textbf{Hard Rain}} & \multicolumn{2}{c}{\textbf{Foggy}} & \multicolumn{2}{c}{\textbf{Foggy Rain}} \\ 
\cmidrule(lr){2-3} \cmidrule(lr){4-5} \cmidrule(lr){6-7} \cmidrule(lr){8-9} \cmidrule(lr){10-11} 
 & \textbf{ASR}      & \textbf{AP}     & \textbf{ASR}       & \textbf{AP}      & \textbf{ASR}        & \textbf{AP}        & \textbf{ASR}    & \textbf{AP}   & \textbf{ASR}          & \textbf{AP}        \\ \midrule
\textbf{Clean}           & 5.56              & 71.44              & 30.56              & 48.41               & 6.94                & 69.17                & 26.39           & 52.98            & 13.89                & 60.46                \\
\textbf{Ours-v2}    & 41.67             & 47.97              & \textbf{79.17}              & 27.73               & \textbf{43.06}               & \textbf{50.09}                & 69.44           & 28.36            & 59.44                & 28.54                \\
\textbf{Ours-v3}    & 29.17             & 50.41              & 74.58              & 33.68               & 34.72               & 52.95                & 63.89           & 34.54            & 62.50                & 33.42                \\
\textbf{Ours-v5}    & 34.58             & 50.81              & 66.67              & 34.98               & 27.78               & 56.54                & 69.44           & 28.33            & 68.06                & 27.72                \\
\textbf{Ours-fr}    & 32.92             & 50.31              & 54.17              & 39.83               & 20.83               & 64.54                & 55.56           & 38.55            & 56.94                & 37.33                \\
\textbf{Ours-ddetr} & \textbf{45.83}             & \textbf{46.20}              & 76.39              & \textbf{27.72}               & 36.11               & 58.33                & \textbf{79.17}           & \textbf{24.68}            & \textbf{79.17}                & \textbf{23.58}                \\ \bottomrule
\end{tabular}
}
\vspace{-4 mm} 
\end{table}

\subsubsection{Complexity Environments Robustness}

\begin{figure*}[t!]
\centering
\includegraphics[width=12 cm]{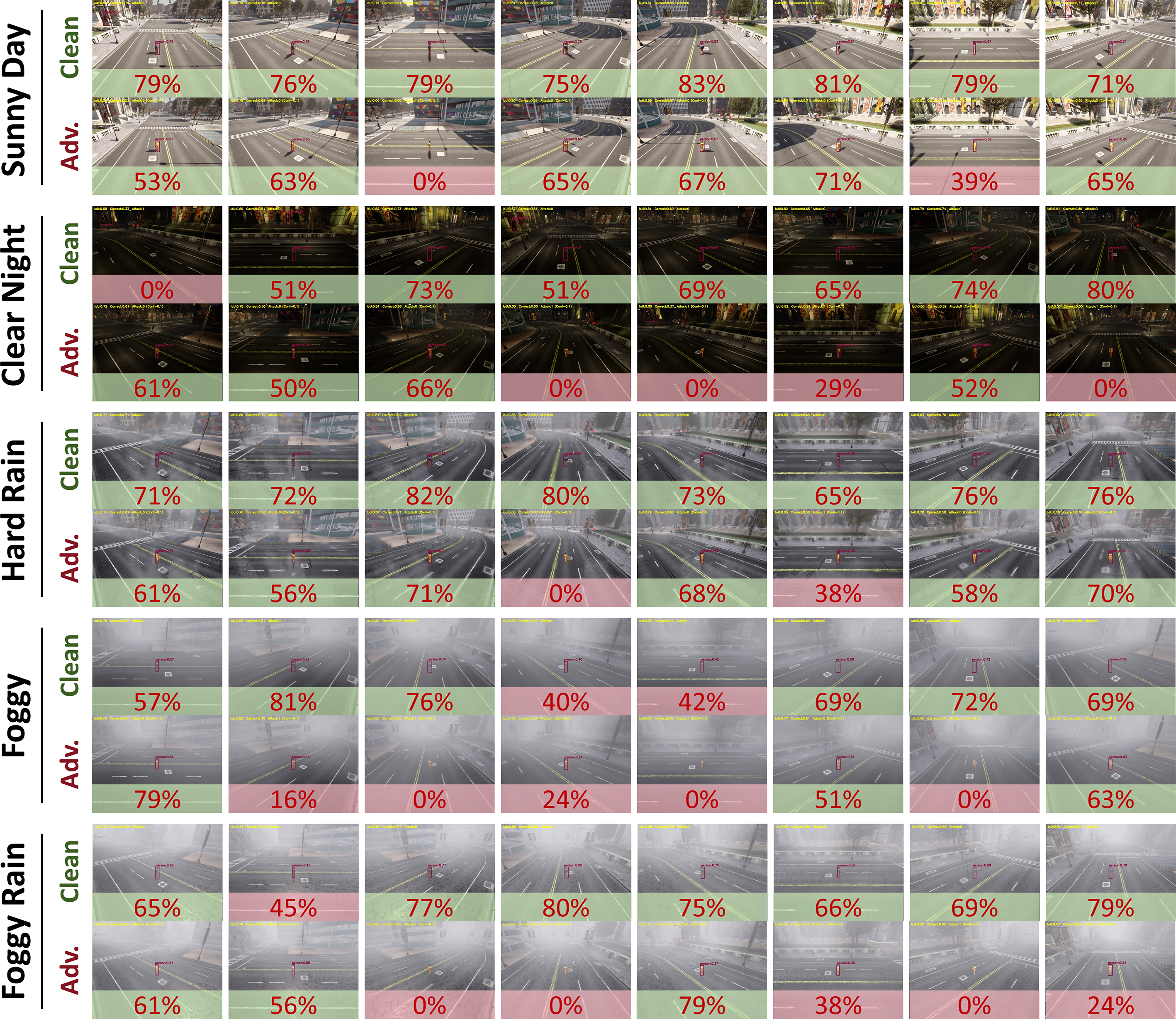}
\caption{Qualitative comparison of detection results under diverse environmental conditions. Each row presents a weather scenario (Sunny Day, Clear Night, Hard Rain, Foggy, Foggy Rain), with clean inputs (top) and AdvSerial adversarial inputs (bottom). Overlaid values indicate YOLO-v5 detection confidence. The adversarial patches consistently suppress detection confidence across conditions, with strongest suppression under low-visibility scenarios.}
\label{Fig6_1}
\vspace{-4 mm} 
\end{figure*}

Generating realistic test scenarios for infrastructure-based perception has attracted increasing attention in intelligent transportation research~\cite{cai2026text2scenario}. To systematically quantify attack stability under complex dynamics, we employ the PADetBench~\cite{lian2025padetbench} framework to construct a comprehensive dataset of 3,600 images per patch. The dataset traverses five distinct weather conditions (\textit{Sunny Day, Clear Night, Hard Rain, Foggy, Foggy Rain}) across varying distances ($8\text{-}12$m) at a fixed pitch angle of $60^\circ$ and eight azimuthal viewpoints ($0^\circ\text{-}315^\circ$), enabling a rigorous assessment against background clutter and weather variations.

The results in Tab.~\ref{weather} demonstrate that AdvSerial maintains robust attack efficacy across a broad spectrum of complex environmental scenarios. We observe a pronounced correlation between weather conditions and attack effectiveness, suggesting that the generated suppression features possess strong semantic saliency. Specifically, under \textit{Foggy Rain} and \textit{Clear Night}, the attack achieves peak ASRs of $79.17\%$ while substantially reducing AP to a range of $23\%$ to $27\%$. Crucially, even under \textit{Sunny Day}, where intense specular reflections prevent a sharp increase in ASR, the attack effectively suppresses detector confidence, reducing AP to approximately $46\%$ from the $71.44\%$ baseline. Qualitative visualization of detection results under each weather condition is provided in Fig.~\ref{Fig6_1}.

\subsubsection{Pitch-angle Robustness}
\label{Pitch_multi_angle}

We evaluate pitch-view robustness using neural-rendered simulations under randomized illuminance ($E_k\sim\mathcal{U}(10,20)\,\mathrm{klx}$). Virtual cameras sample pitch angles uniformly from $0^\circ$ to $180^\circ$ at a $10\,\text{m}$ radius. We aggregate the frame-level ASR into $10^\circ$ bins.

\begin{figure*}[t!]
\centering
\includegraphics[width=12 cm]{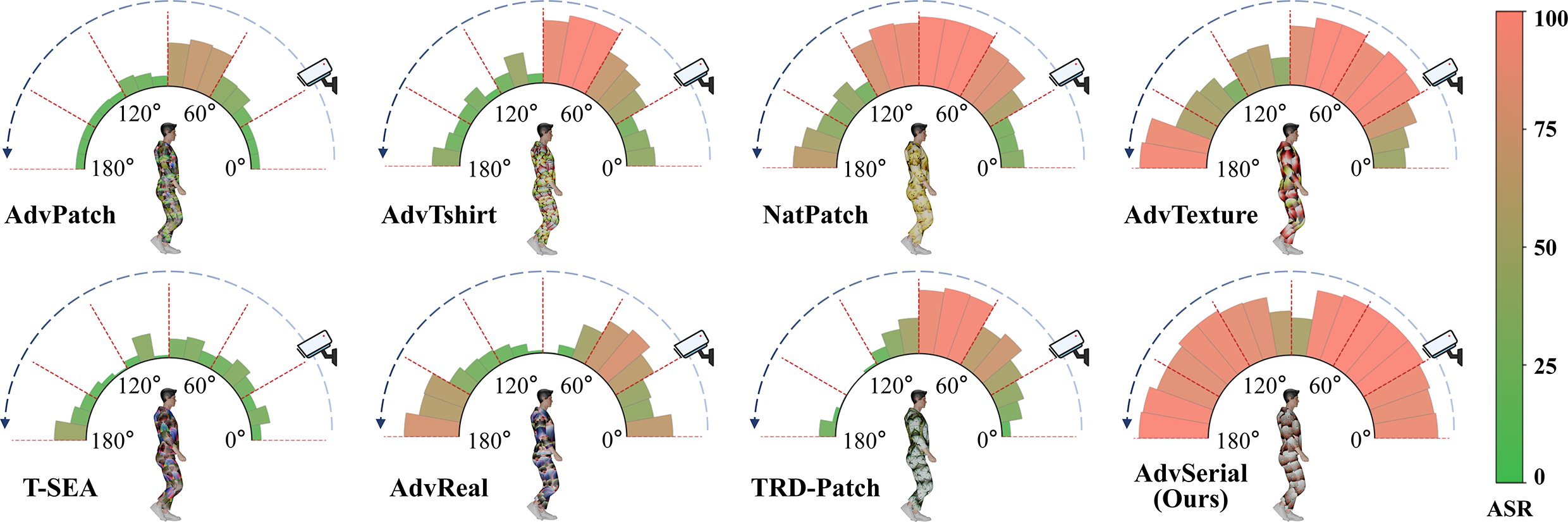}
\caption{Pitch angle robustness evaluation in simulation. Comparison of frame-level ASR across viewing angles ranging from $0^\circ$ to $180^\circ$ for different attack methods.}
\label{pitch_attack}
\vspace{-4 mm} 
\end{figure*}

As shown in Fig.~\ref{pitch_attack}, most baselines suffer severe ASR degradation beyond $60^\circ$, particularly in the rear hemisphere ($90^\circ$--$180^\circ$). Within the $60^\circ$--$90^\circ$ interval, AdvSerial shows a noticeable ASR drop, declining to approximately $57.3\%$ in the steepest $80^\circ$--$90^\circ$ bin. Specifically, in this narrow band, global texture-based methods (e.g., AdvTexture, NatPatch) temporarily outperform our approach by maintaining ASRs above $\sim 75\%$. This decline stems from extreme foreshortening and self-occlusion at near-vertical angles. While uniform textures remain visible on the shoulders, the head and upper body partially occlude AdvSerial's core disruptive features on the torso. Nevertheless, AdvSerial's superior overall robustness outweighs this localized vulnerability. It excels in standard surveillance views ($0^\circ$--$60^\circ$, ASR $\sim 100\%$). Furthermore, it maintains near-perfect ASR ($\ge 95\%$) across the entire $90^\circ$--$180^\circ$ rear spectrum, where all other baselines' attack performance significantly decreased.

\subsubsection{Defense Robustness}
\label{sec:defense}
\textbf{Defense based on NapGuard.}
\label{sec:defense_sparse4d}
To assess the robustness against defenses, we evaluate detectability using NapGuard~\cite{Wu_2024_CVPR} on $525$ real-world images (Tab.~\ref{tab_detect}). Methods relying on standard tiling or unconstrained optimization, such as T-SEA and AdvReal, introduce sharp cross-seam feature discontinuities that trigger anomaly detectors, resulting in over $350$ flagged frames with high confidence ($>0.66$). Conversely, naturalistic designs (e.g., NatPatch, TRDPatch) demonstrate improved stealth but often at the cost of reduced attack potency. 

AdvSerial achieves an optimal balance, triggering only a single detection ($1/525$). This exceptional evasiveness serves as a macroscopic empirical validation of the bound-tightening property formulated in Eq.~\ref{eq:tighter_bound}. NapGuard explicitly targets high-frequency feature artifacts; by routing seams through low-sensitivity regions, FSQ actively minimizes the worst-case cross-seam feature bound $\Phi_\ell(\mathcal{S})$ (Eq.~\ref{eq:phi_bound}). Consequently, the actual feature discontinuity $\|\Delta F_\ell(p)\|$ is strictly constrained. This structural constraint prevents the localized feature spikes that typically expose adversarial patches to boundary-aware defenses, thereby retaining strong attack performance while maintaining spatial continuity.

\begin{table}[t!]
\renewcommand{\arraystretch}{1} 
\centering
\caption{Detectability evaluation against NapGuard. Comparison of detection frequency and confidence scores for different adversarial patches.}
\label{tab_detect}
\resizebox{8.6cm}{!}{
\begin{tabular}{cccccc}
\hline
\multicolumn{1}{c}{Method} & \multicolumn{1}{c}{Conf.} & \multicolumn{1}{c}{Det. Num.} &
\multicolumn{1}{c}{Method} & \multicolumn{1}{c}{Conf.} & \multicolumn{1}{c}{Det. Num.} \\
\hline
AdvPatch   & 0.61                      & 133/525 &
TRDPatch   & \cellcolor{lightgreen}0       & \cellcolor{lightgreen}0 \\
AdvTshirt  & \cellcolor{lightyellow}0.63 & \cellcolor{lightyellow}24/525 &
T-SEA      & 0.67                      & 378/525 \\
AdvTexture & \cellcolor{lightgreen}0.59  & \cellcolor{lightyellow}47/525 &
AdvReal    & 0.67                       & 352/525\\
NatPatch   & \cellcolor{lightgreen}0.58  & \cellcolor{lightgreen}5/525 &
AdvSerial (ours) & \cellcolor{lightyellow}0.61 & \cellcolor{lightgreen}1/525   \\
\hline
\end{tabular}
}
\vspace{-4 mm}
\end{table}

\begin{table*}[ht!]
\renewcommand{\arraystretch}{1.1} 
\centering
\caption{Evaluation against the 3D-temporal detector Sparse4D-v3. Comparison of ASR and AP on the nuScenes-mini dataset under sequence-level defense.}
\label{tab_3d_defense}
\resizebox{14cm}{!}{ 
\begin{tabular}{lccccccccc}
\hline
\multicolumn{1}{c}{\textbf{Metric}} & \textbf{Clean} & \textbf{AdvPatch} & \textbf{AdvTshirt} & \textbf{NatPatch} & \textbf{T-SEA} & \textbf{AdvTexture} & \textbf{TRD-Patch} & \textbf{AdvReal} & Ours \\
\hline
\textbf{ASR (\%) $\uparrow$} & 36.01 & 40.53 & 64.61 & 57.82 & 36.21 & 42.59 & 41.36 & 36.21 & \textbf{65.84} \\
\textbf{AP (\%) $\downarrow$}  & 59.97 & 60.73 & 54.51 & 53.34 & 57.64 & 56.82 & 61.01 & 57.60 & \textbf{51.91} \\
\hline
\end{tabular}
}
\end{table*}

\begin{figure}[t!]
\centering
\includegraphics[width=9 cm]{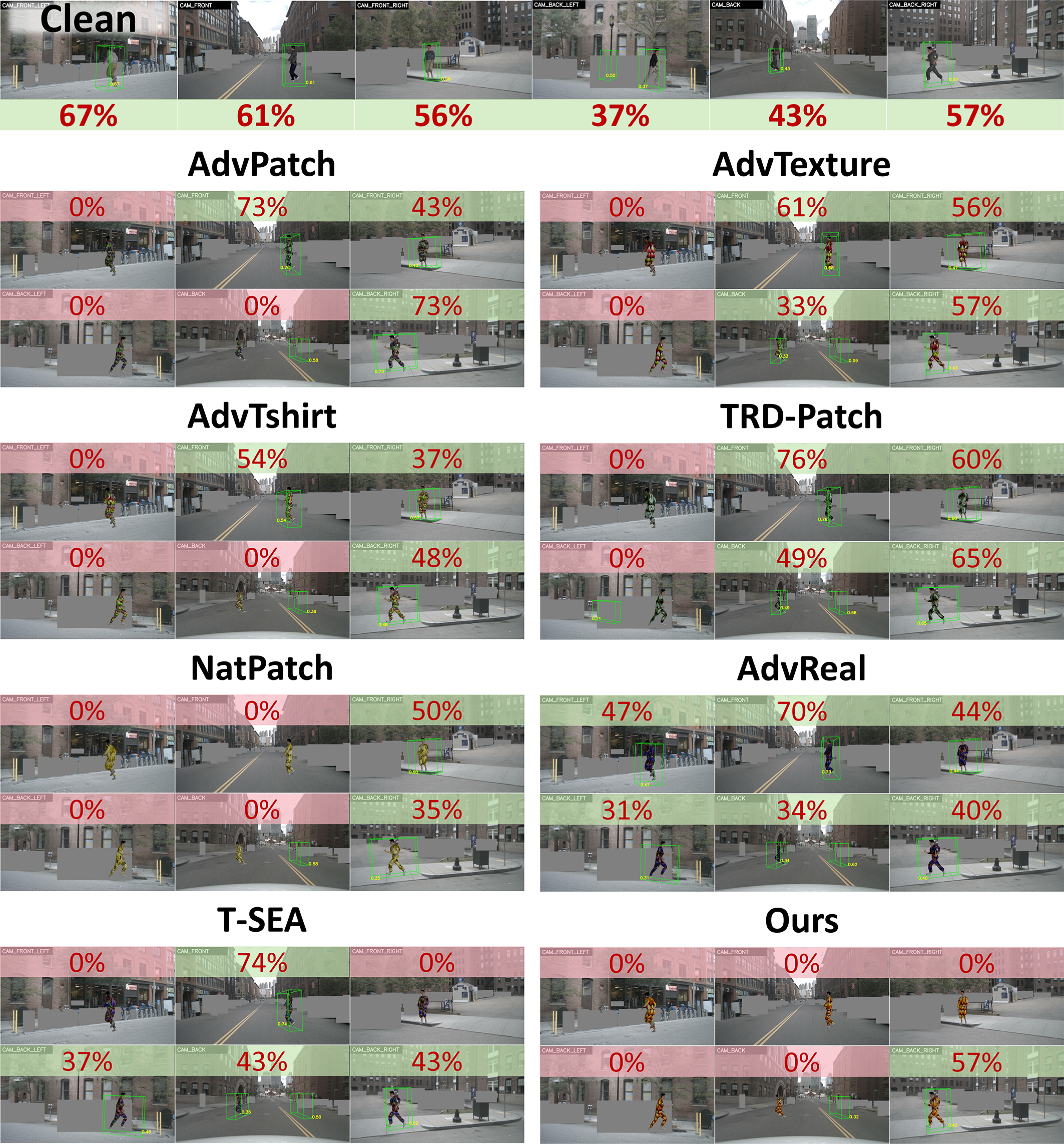}
\caption{Qualitative results of adversarial attacks against Sparse4D-v3.}
\label{Fig6_2}
\vspace{-4 mm} 
\end{figure}

\textbf{Defense based on 3D-Temporal Perception (Sparse4d-v3).}
To further verify robustness against "continuous" defenses relevant to autonomous driving, we extend our evaluation to Sparse4D-v3~\cite{lin2023sparse4d}, a SOTA 3D detector employs temporal fusion and sparse query mechanisms. Experiments are conducted on the nuScenes-v1.0mini dataset. As shown in Tab.~\ref{tab_3d_defense}, existing 2D-based SOTA methods struggle to transfer to this 3D pipeline. T-SEA and AdvReal yield ASRs ($\approx 36.2\%$) almost identical to the Clean baseline ($36.01\%$), indicating a complete failure. Their inconsistent frame-level perturbations are effectively filtered out by the detector's temporal aggregation module. In contrast, AdvSerial effectively disrupts the pipeline, significantly boosting the ASR to $65.84\%$ and degrading AP to $51.91\%$. By achieving consistent global suppression across multi-view sensors, AdvSerial prevents valid sparse query initialization, thereby overcoming the robustness of temporal fusion. Qualitative detection results on representative nuScenes sequences are provided in Fig.~\ref{Fig6_2}.

For the defense evaluation against Sparse4D-v3, we follow its standard benchmark protocol on the \href{https://www.nuscenes.org/}{nuScenes-v1.0mini} dataset, as described in Sec.~\ref{sec:defense_sparse4d}. 


\subsection{Physical World Experiments}
\label{sec:physical}

Consistent with our digital evaluations, all physical world experiments strictly employ a confidence threshold of 0.5 and an Intersection over Union (IoU) threshold of 0.5. We conduct physical experiments in both indoor and outdoor environments to ensure comprehensive evaluation. We mount the indoor cameras at a height of 3.0 meters. We install the outdoor cameras at an elevated height of 5.0 meters. These dual configurations realistically simulate the steep downward pitch angles typical of diverse infrastructure surveillance systems. We perform the distance robustness and full-azimuth angle evaluations exclusively in the 5.0-meter outdoor setting. We conduct the postural sensitivity assessments across both indoor and outdoor environments. Our preliminary analysis reveals no significant difference in the attack success rate between the indoor and outdoor scenarios. Consequently, we directly average the results from all corresponding test sets to compute the final reported metrics. A detection is counted as a failure if it yields an incorrect category, a confidence score $\le 0.5$, or an IoU $\le 0.5$.


\subsubsection{Distance Robustness Study}

\begin{figure*}[t!]
\centering
\includegraphics[width=13 cm]{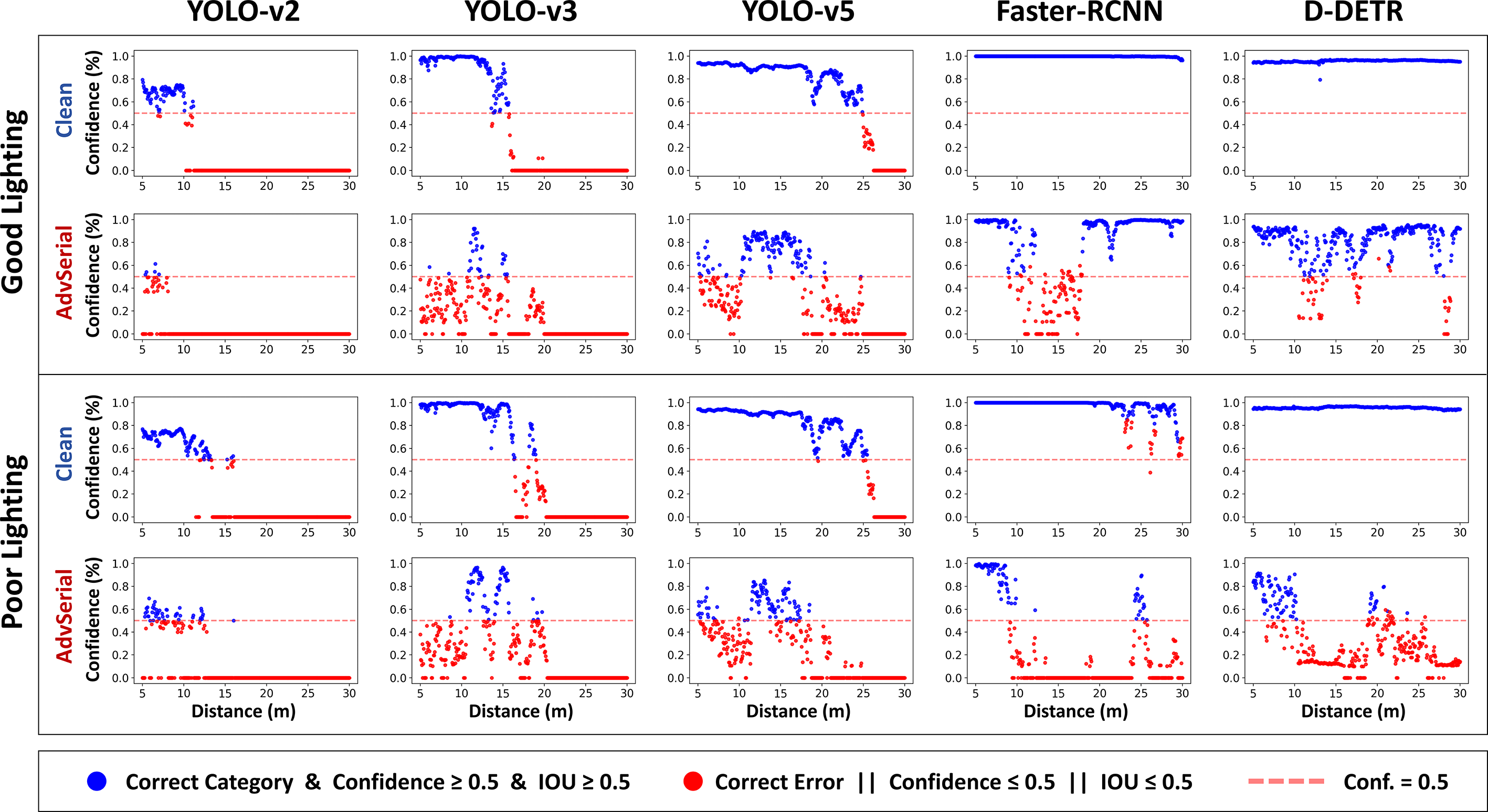}
\caption{Physical world evaluation under varying distances and lighting. Scatter plots show detection confidence and correctness for Clean vs. AdvSerial samples as the distance increases from 5m to 30m.}
\label{fig:distance}
\vspace{-4 mm} 
\end{figure*}

We evaluate distance robustness by recording sequences where the subject walks from $5$ to $30\,\text{m}$ along the optical axis under good and poor lighting. A detection is counted as a failure if the prediction is incorrect, confidence $\le 0.5$, or IoU $\le 0.5$. Results are shown in Fig.~\ref{fig:distance}.

Under good lighting, clean detectors remain stable across their effective ranges. With AdvSerial, confidence and localization steadily deteriorate: YOLO-v2/v3 stay mostly below $0.5$ confidence; YOLO-v5 yields only sporadic correct detections; Faster R-CNN and D-DETR degrade at medium-to-long distances while remaining more reliable at close range. Under poor lighting, clean models lose $2$--$5\,\text{m}$ of stable range, and AdvSerial pushes the failure boundary even closer, especially for the YOLO family.

The trends align with the underlying mechanism. As distance grows, pixel coverage and feature signal-to-noise ratio drop, making one-stage detectors more susceptible to objectness suppression. Two-stage and transformer models benefit from RPN proposals and self-attention at short range but also degrade at long range due to lower proposal scores and attention diffusion.


\begin{figure}[t!]
\centering
\includegraphics[width=7 cm]{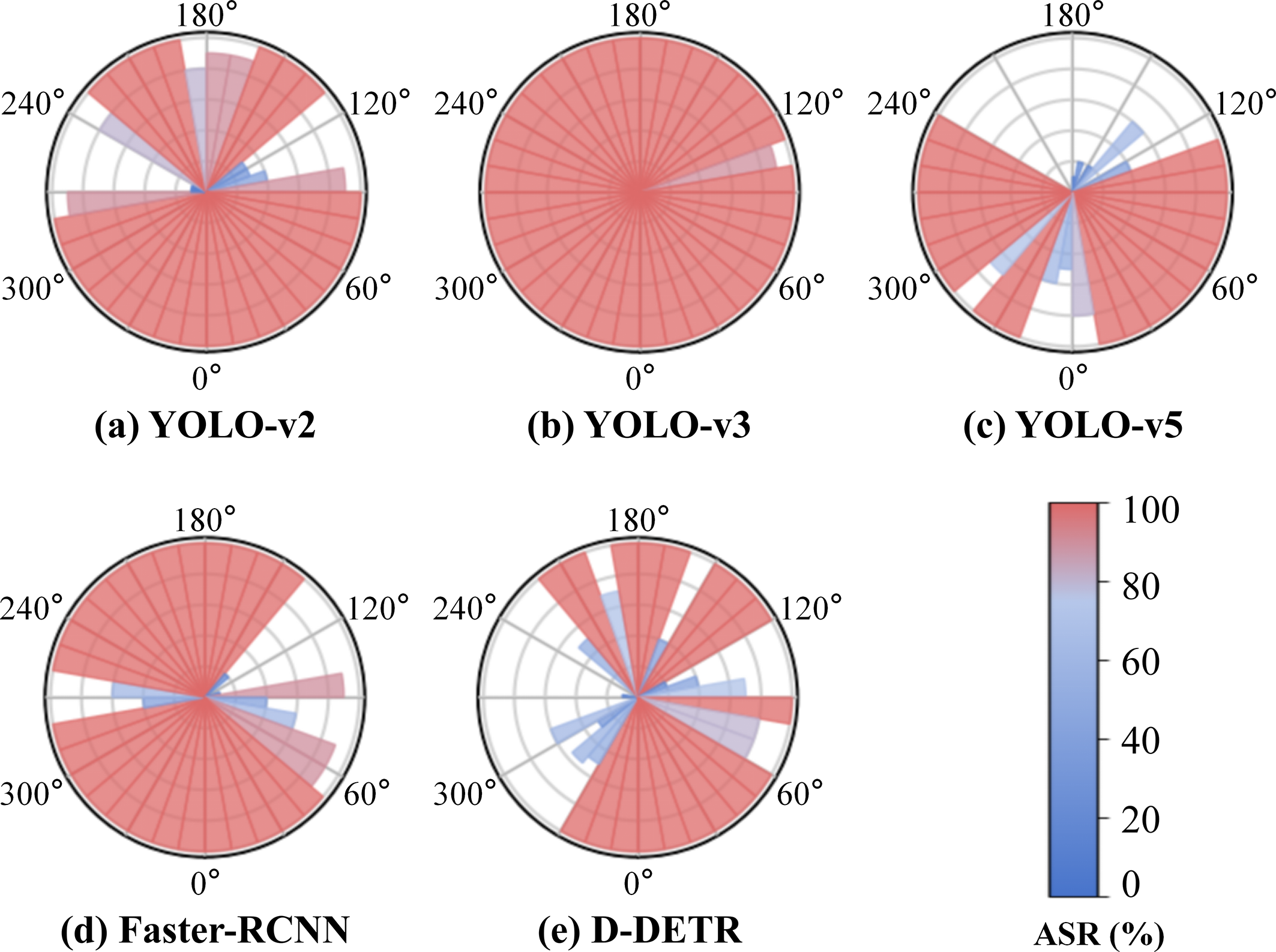}
\caption{Full-azimuth attack performance in the physical world. Polar plots illustrate the ASR distribution across $360^\circ$ viewing angles for five different detectors.}
\label{lidar_attack}
\vspace{-4 mm} 
\end{figure}

\subsubsection{Full-azimuth Robustness Study}

We assess full-azimuth robustness of proposed patches in the physical world. Similar to Sec.~\ref{Pitch_multi_angle}, we fix the camera subject distance at 3 m and average ASR of angle ranges over 10 set of experiments. 

Fig.~\ref{lidar_attack} illustrates azimuthal ASR distributions for patches adversarially generated by AdvSerial across five detectors: YOLO-v2, YOLO-v3, YOLO-v5, Faster R-CNN, and D-DETR. The results show that all patches exhibit uneven ASR distributions, revealing clear directional sensitivity in adversarial effectiveness. For YOLO-v2 and YOLO-v3 in Fig.~\ref{lidar_attack} (a) and (b), the ASR remains high across most azimuths but sharply degrades in a few rear-view sectors, where ASR drops close to zero. In contrast, Fig.~\ref{lidar_attack} (c) (YOLO-v5) demonstrates greater attacking consistency, achieving over 80\% ASR in both frontal and lateral directions, although degradation appears at certain rear angles.

\subsubsection{Postural Sensitivity}

\begin{figure*}[t!]
\centering
\includegraphics[width=16 cm]{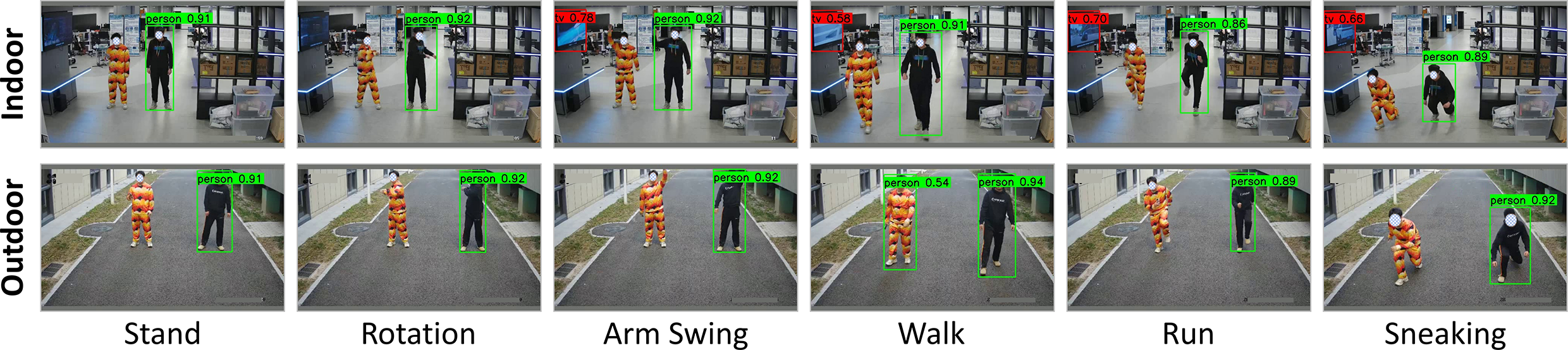}
\caption{Qualitative results in physical scenarios. Visualization of attack effectiveness across diverse postures and scenes (Indoor/Outdoor).}
\label{physical}
\vspace{-4 mm} 
\end{figure*}

\begin{figure}[t!]
\centering
\includegraphics[width=8.6 cm]{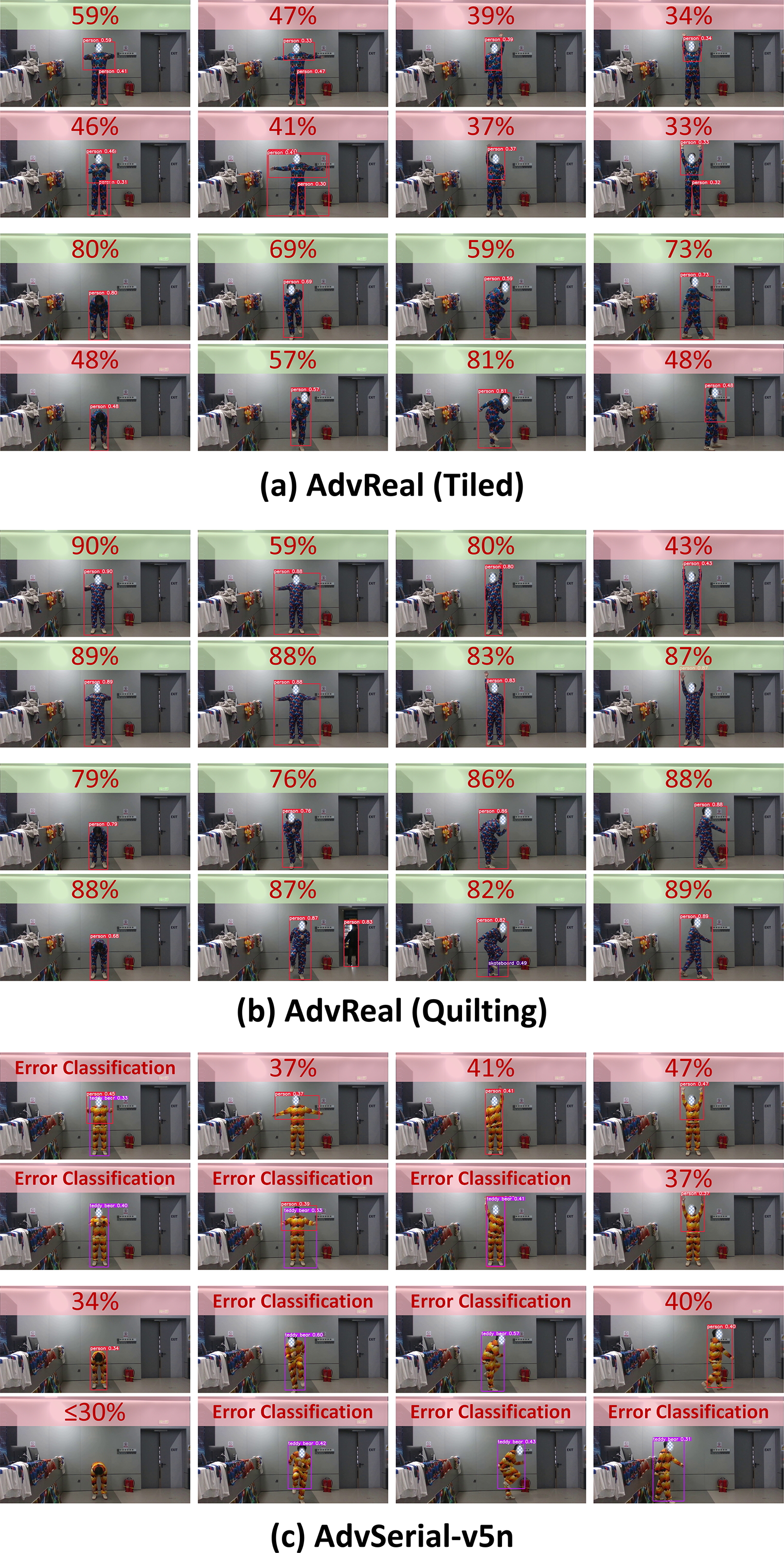}
\caption{Comparison of attack mechanisms under quilting, both the black-box and white-box models are YOLO-v5n.}
\label{Fig11_1}
\vspace{-4 mm} 
\end{figure}

\begin{table*}[t]
\renewcommand{\arraystretch}{1}
\centering
\caption{Physical robustness evaluation across diverse human postures. Comparison of Detection Rate (\%) and Average Confidence scores between 5 different samples across six distinct motion states (e.g., Stand, Walk, Run)}
\label{posture_clean_advserial}
\resizebox{13cm}{!}{
\begin{tabular}{lcccccc}
\hline
\multirow{2}{*}{\textbf{Setting}} &
\textbf{Stand} & \textbf{Rotation} & \textbf{Arm-swing} & \textbf{Walk} & \textbf{Run} & \textbf{Sneaking} \\
\cline{2-7}
 & \multicolumn{6}{c}{Detection Rate (\%) / Confidence} \\
\hline
\textbf{Clean}     & 100.00 / 0.92 & 100.00 / 0.93 & 100.00 / 0.91 & 100.00 / 0.90 & 100.00 / 0.89 & 100.00 / 0.92 \\
\textbf{NatPatch}  & 75.42 / 0.71 & 79.18 / 0.74 & 88.65 / 0.81 & 85.34 / 0.78 & 82.51 / 0.75 & 72.86 / 0.68 \\
\textbf{AdvTexture}& \textbf{8.42} / 0.23 & 21.67 / 0.37 & 61.34 / 0.55 & 53.28 / 0.57 & 48.93 / 0.48 & 16.75 / 0.27 \\
\textbf{AdvReal}   & 14.36 / 0.27 & 23.41 / 0.32 & 52.18 / 0.50 & 57.35 / 0.53 & 41.76 / 0.44 & 12.48 / 0.25 \\
\textbf{AdvSerial} & 8.73 / \textbf{0.21} & \textbf{17.58} / \textbf{0.25} & \textbf{29.27} / \textbf{0.34} & \textbf{35.63} / \textbf{0.39} & \textbf{15.82} / \textbf{0.26} & \textbf{7.54} / \textbf{0.18} \\
\hline
\end{tabular}
}
\vspace{-4mm}
\end{table*}

We record $10$ physical-world videos in total ($5$ indoor and $5$ outdoor), each containing a full sequence of postures: standing, rotation, arm swing, walking, running, and sneaking. The adversarial texture is trained against YOLO-v5n, and both confidence and IoU thresholds are standardized at $0.5$. As shown in Fig.~\ref{physical}, the adversarial texture remains visually effective across diverse poses and scenes. Quantitatively, Tab.~\ref{posture_clean_advserial} confirms that clean clothing yields near-perfect detection, with a detection rate close to $100\%$ and an average confidence of approximately $0.91$ across all postures. NatPatch, which prioritizes visual naturalness over attack strength, reduces the detection rate to $72\%\sim89\%$ across postures, reflecting the trade-off between perceptual concealment and physical-world suppression efficacy.

After applying the AdvSerial texture, both detection rate and confidence drop markedly. In relatively static or moderate poses (standing, rotation, running, and sneaking), detection rates fall to $7\%\sim18\%$ with confidence scores around $0.18\sim0.26$, indicating stable suppression under conditions representative of typical pedestrian presence in infrastructure monitoring scenarios. However, arm swing and walking remain challenging, with detection rates rising to $29\%\sim36\%$. Highly articulated motions cause severe self-occlusion and break the spatial continuity of the adversarial pattern, physically blocking effective attack features. Meanwhile, dynamic poses maximize the exposure of robust, unperturbed somatic features, providing the detector with structural information that overrides the partial texture suppression. AdvTexture reduces the detection rate to $8\%\sim21\%$ under static poses yet degrades substantially under dynamic motions, with detection rates reaching $53\%\sim61\%$ for arm swing and walking, whereas AdvSerial maintains $29\%\sim36\%$ under the same conditions. Compared with AdvReal, the most recent baseline optimized for physical robustness, AdvSerial achieves lower detection rates across all postures, with AdvReal degrading to $52\%\sim57\%$ under dynamic motions. AdvSerial maintains $29\%\sim36\%$, benefiting from the serial-frame loss that regularizes temporal consistency across pose transitions. From an infrastructure deployment perspective, even the $29\%\sim36\%$ detection rate under the most challenging motion conditions still represents a $64\%\sim71\%$ reduction from normal operation, posing a significant security risk to surveillance systems.

\begin{figure*}[t!]
\centering
\includegraphics[width=16 cm]{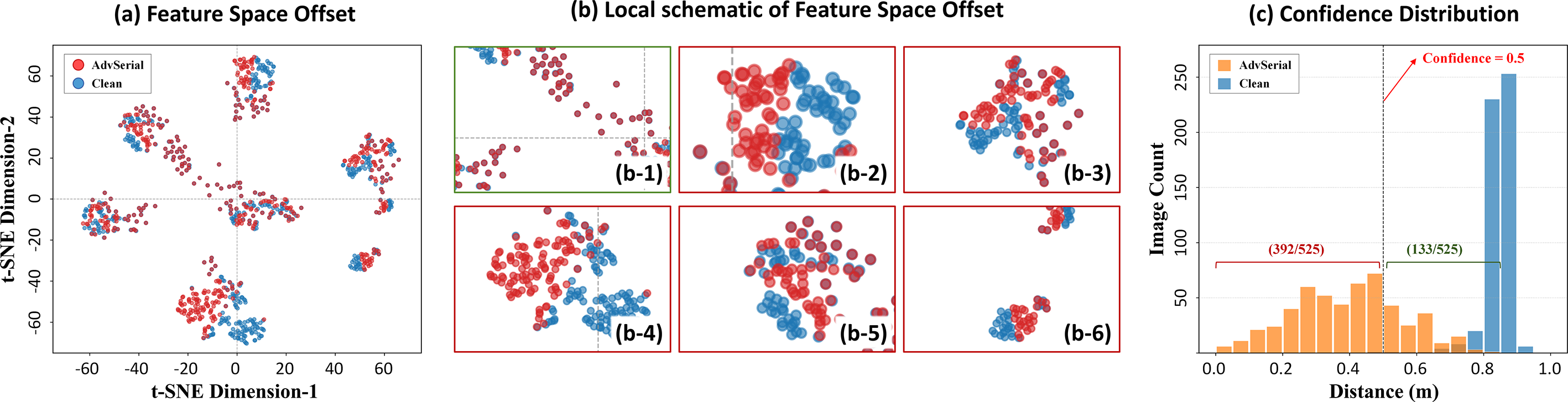}
\caption{Feature space analysis of AdvSerial on YOLO-v5n. (a) Global t-SNE projection of person features for clean (blue) and AdvSerial (red) samples, showing systematic feature displacement. (b) Six local views (b-1 through b-6) illustrating varying degrees of separation between clean and adversarial representations within scene clusters. (c) Per-image person confidence distribution, where 392/525 adversarial images fall below the 0.5 threshold.}
\label{Figure16}
\end{figure*}

To further illustrate the robustness of the suppression-centric mechanism, Fig.~\ref{Fig11_1} compares the quilting resilience of AdvReal and AdvSerial across postures. AdvReal with direct tiling retains its activation-guided attack efficacy. However, after quilting, AdvReal loses effectiveness because the boundary cut disrupts the pixel distributions required for feature redirection. In contrast, AdvSerial after quilting maintains robust suppression across all postures, with person confidence reduced below the detection threshold or misclassified as non-person categories. The comparison confirms that suppression-centric feature erasure is inherently resilient to quilting, whereas activation-guided attacks depend on fine-grained pixel alignment that quilting disrupts.


\subsection{Explainability}
\label{sec:explainability}

\subsubsection{Interpretability via t-SNE}

To visualize how the adversarial patch reshapes the detector's internal representations, we collect $1{,}050$ real-world images ($525$ with AdvSerial-v5 texture, $525$ clean), extract multi-scale features before the YOLO-v5 Detect head, reduce them with PCA, and project them into two dimensions via t-SNE. Because t-SNE preserves local neighborhood structure, we can examine how adversarial inputs shift within scene clusters and relate these shifts to changes in detection confidence.

Fig.~\ref{Figure16}(a) shows the global t-SNE map. The two distributions occupy partially overlapping but systematically displaced regions, indicating that the patch induces a consistent directional shift in person representations rather than random scattering. Fig.~\ref{Figure16}(b) provides six representative local views (b-1 through b-6) revealing varying degrees of displacement. In some regions (e.g., b-1, b-4), clean and adversarial points nearly overlap, suggesting weak feature displacement in certain scene contexts. In others (e.g., b-2, b-3, b-5, b-6), the two groups are clearly separated within the same scene cluster, evidencing systematic migration of pedestrian features and weakened compactness of person-related representations.

Fig.~\ref{Figure16}(c) reports the per-image person confidence distribution. Clean samples cluster overwhelmingly above $0.8$, whereas adversarial samples shift toward substantially lower values spanning $0.0$ to $0.6$. With a detection threshold of $0.5$, $392$ of $525$ adversarial images fall below the boundary while $133$ remain above, yielding an ASR of $74.8\%$ and reducing mean person confidence from $84.30\%$ to $39.38\%$. These results confirm that the patch suppresses person-specific semantic feature activations in mid- and high-level layers, attenuates the person logits, and drives the majority of samples below the detection threshold.

\begin{figure*}[t!]
\centering
\includegraphics[width=16 cm]{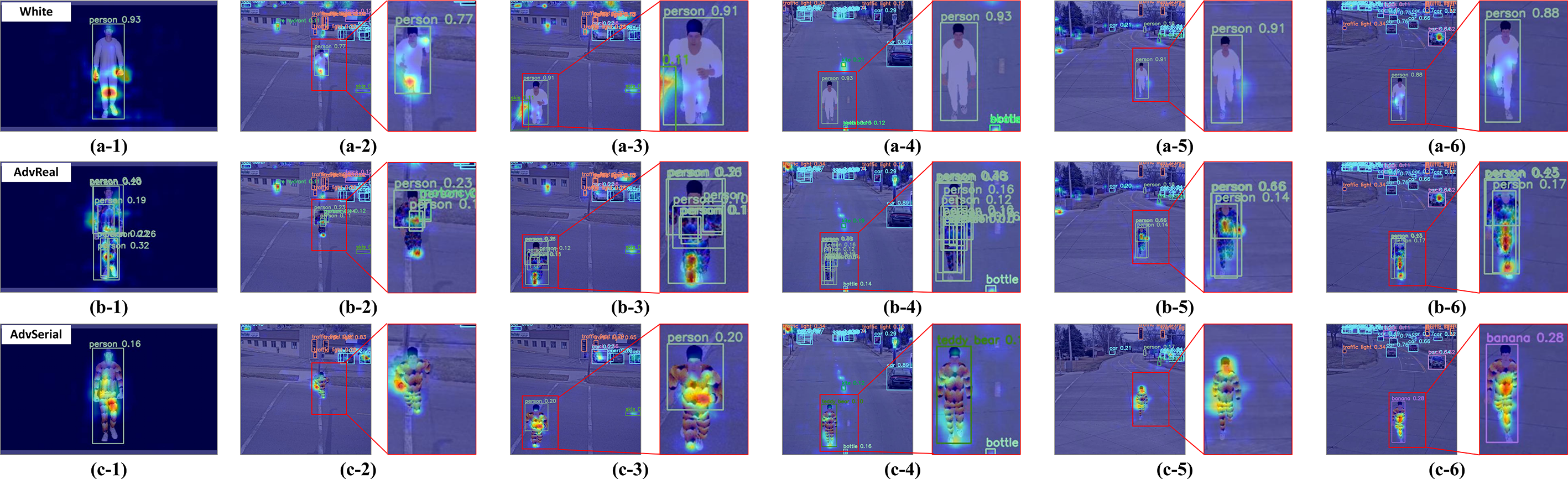}
\caption{Grad-CAM visualization. Comparison of attention heatmaps for Clean, AdvReal, and AdvSerial, highlighting differences in detector focus regions.}
\label{heatmap}
\vspace{-6 mm} 
\end{figure*}


\begin{table}[t!]
\centering
\caption{Quantitative analysis of seam-aware feature activations on YOLO-v2. Comparison of seam gradient sensitivity (Seam $w$), cross-seam feature discontinuity ($\|\Delta F\|$), and global feature magnitude ($\|F\|$).}
\label{tab_mechanism}
\resizebox{8.2cm}{!}{
\begin{tabular}{lccc}
\toprule
\textbf{Method} & \textbf{Seam $w \downarrow$} & \textbf{Mean $\|\Delta F\| \downarrow$} & \textbf{Mean $\|F\| \downarrow$} \\
\midrule
T-SEA      & 0.140 & 0.775 & 3.424 \\
AdvTexture & 0.165 & 0.857 & 3.508 \\
AdvReal    & 0.109 & 0.727 & 3.627 \\
\textbf{AdvSerial (Ours)} & \textbf{0.083} & \textbf{0.558} & \textbf{3.207} \\
\bottomrule
\end{tabular}
}
\vspace{-4 mm} 
\end{table}

\subsubsection{Visualization and Mechanism Analysis}
\label{sec_mechanism}

To explicitly visualize the attack mechanisms and empirically validate the bound tightening property of our Feature Smooth Quilting (FSQ), we conduct a joint quantitative and qualitative feature activation analysis. Tab.~\ref{tab_mechanism} details the microscopic feature dynamics of representative methods, while Fig.~\ref{heatmap} visualizes the GradCAM~\cite{selvaraju2017grad} attention heatmaps across six representative samples under three settings.

\textbf{Local Pathfinding and Feature Smoothness.}
As reported in Tab.~\ref{tab_mechanism}, existing methods struggle with boundary feature continuity regardless of their tiling strategies. T-SEA, which represents naive direct splicing, yields high seam gradient sensitivity (Seam $w = 0.140$) and severe feature discontinuity ($\|\Delta F\| = 0.775$). AdvTexture attempts to resolve this through scalable boundary-free generative patterns, yet under complex 3D physical mapping it still suffers from extreme sensitivity peaks at wrap-around seams (Seam $w = 0.165$) and pronounced feature fractures ($\|\Delta F\| = 0.857$), revealing a critical gap between visual continuity and deep feature continuity. Even AdvReal, which heavily optimizes for physical realism, traverses sensitive boundaries with Seam $w = 0.109$, resulting in distinct feature discontinuities ($\|\Delta F\| = 0.727$).

In contrast, FSQ actively routes the seam through low-sensitivity channels in the loss landscape, reducing Seam $w$ to $0.083$. This intelligent pathfinding directly translates to microscopic feature smoothness, bounding the cross-seam feature discontinuity $\|\Delta F\|$ to $0.558$, a reduction of $23.2\%$ relative to AdvReal. This quantitative reduction provides direct empirical evidence supporting the theoretical bound in Eq.~\ref{eq:tighter_bound}, confirming that minimizing the feature-weighted pixel error defined in Eq.~\ref{eq:weighted_error} effectively eliminates sharp deep feature artifacts at tile boundaries.

\textbf{Global Semantic Suppression.}
This localized boundary smoothness is the prerequisite for achieving global semantic feature suppression, which is reflected in the global feature magnitude $\|F\|$. A lower $\|F\|$ indicates that the texture has absorbed and dampened the overall feature energy available to the detector for person classification. Traditional methods such as AdvReal inject high-frequency visual boundaries during quilting, which act as non-discriminative artifacts that trigger false activation peaks, sustaining a high global feature magnitude ($\|F\| = 3.627$). By eliminating these artificial boundaries via FSQ, AdvSerial prevents the injection of spurious activations and reduces the global feature magnitude to $3.207$.

This suppression mechanism is directly confirmed by the GradCAM maps in Fig.~\ref{heatmap}. Under the clean setting (row a), heatmaps display diffuse but semantically coherent responses along the canonical head, torso, and limb structure, with high detection confidence across all six samples. Under AdvReal (row b), activation concentrates into localized high-intensity saliency peaks on the patch region, reflecting an attention-hijacking strategy that partially degrades detection confidence yet still preserves residual person-structure responses. Under AdvSerial (row c), the activation pattern changes fundamentally. The heatmaps become globally diffuse and detached from the human body configuration, with no identifiable structural anchoring, and in several samples the detector misclassifies the subject as non-person categories. This confirms that AdvSerial operates through a suppression-oriented paradigm rather than attention redirection, dismantling mid- and high-level person features without generating the anomalous high-frequency boundary spikes that typically expose physical patches to boundary-aware defenses.





\section{Discussion and Limitations}
\label{discussion}

\subsection{Mechanism Analysis and Security Implications}

\textbf{Mechanism Analysis.} AdvSerial prioritizes temporal consistency and cross-view generalization over strategies that employ hybrid objectives. Existing attacks jointly target confidence, localization, and category confusion. However, such optimization requires delicate texture alignment that physical distortions frequently disrupt, resulting in transient instability from which downstream models can recover. In contrast, the suppression-centric strategy directs optimization toward fundamental feature erasure, preventing gradient conflicts and ensuring superior physical robustness. The mechanism is confirmed in Fig.~\ref{Fig11_1}. AdvReal's activation-guided pattern retains efficacy with direct tiling (Fig.~\ref{Fig11_1}(a)) but fails after quilting (Fig.~\ref{Fig11_1}(b)) because the boundary cut disrupts the pixel distributions required for feature redirection. AdvSerial remains effective after quilting (Fig.~\ref{Fig11_1}(c)) since feature erasure does not depend on fine-grained pixel alignment. Across diverse one- and two-stage detectors, including anchor-based and anchor-free architectures, AdvSerial attains an ASR up to $89.7\%$ in simulation and sustains high efficacy at ranges up to $40\mathrm{m}$, with an average maximum consecutive-frame loss of approximately $11$ frames and a mean person confidence of $0.22$.

The durable suppression enables AdvSerial to bypass boundary-based defenses such as NapGuard and to penetrate sequence-level perception pipelines. State-of-the-art defenses detect physical attacks by exploiting flickering or spatiotemporal inconsistency~\cite{man2023person}, yet the Serial Frame Loss reshapes the loss landscape into a flat minimum, maintaining stable invisibility without triggering trajectory-aware anomaly checks. Continuous feature loss effectively contaminates the temporal buffer, as Fig.~S6 confirms under arm swing with maximal body-feature exposure across four viewing directions.

\textbf{Security Implications} The findings carry direct implications for the design and operation of AI-based infrastructure surveillance systems. Modern transportation infrastructure increasingly relies on visual perception for pedestrian detection, traffic monitoring, and incident response, yet the physical adversarial vulnerability of these systems has received little systematic attention from an engineering deployment perspective. AdvSerial reveals that infrastructure-mounted detectors remain vulnerable to physically realizable adversarial textures even when equipped with temporal fusion or boundary-aware defenses. The results suggest several actionable guidelines for infrastructure operators. Cameras mounted at steep downward angles exceeding $45^\circ$ provide natural geometric resistance to torso-based adversarial patches, as near-vertical viewpoints partially occlude the adversarial texture region. Multi-camera configurations with overlapping fields of view can compensate for single-camera suppression failures by providing redundant detection coverage from diverse azimuths. Temporal fusion implemented at the system level, rather than within individual detectors, offers stronger resistance to sustained frame-level attacks by aggregating evidence across longer horizons. Beyond these near-term measures, robust long-term deployment requires defense mechanisms that incorporate motion-aware modeling, explicit treatment of 3D clothing deformation, and cross-camera spatiotemporal consistency verification. AdvSerial thus serves both as an attack generator and as a diagnostic benchmark for assessing and improving the security readiness of AI-enabled civil infrastructure perception systems.

\subsection{Limitations and Future Work}

While the experimental outcomes are robust, certain intrinsic constraints warrant discussion.

\textbf{Suppression-Centric Attack Mechanism.} The current framework prioritizes suppressing person-specific feature activations and detection confidence over inducing targeted misclassifications. The design effectively exploits a specific failure mode by driving detection scores beneath a predefined threshold. However, the framework remains potentially vulnerable to systems employing aggressive temporal smoothing or multi-frame fusion. Future work will explore hybrid optimization objectives that simultaneously manipulate confidence, localization, and category logits while preserving the temporal consistency property that distinguishes AdvSerial from prior attacks.

\textbf{Pose and Deformation Sensitivity.} Despite the integration of 3D motion modeling, Tab.~\ref{posture_clean_advserial} reveals performance variance across postures. Static or mildly articulated poses suffer significant suppression with detection rates below $18\%$, whereas high-amplitude motions such as walking maintain higher detection rates of $29\%\text{-}36\%$. The disparity likely arises from severe self-occlusion and exposure of unperturbed somatic features. From an infrastructure security perspective, the residual detection rate still represents a $64\%\text{-}71\%$ reduction from normal operation and remains a substantial risk. Future research will incorporate high-amplitude motion priors and deformation-aware texture mapping into the 3D training pipeline to close this gap.

\textbf{Limited Defense and Deployment Coverage.} The defense evaluation, currently restricted to NapGuard and Sparse4D-v3, yields encouraging robustness results but primarily reflects performance against a specific class of patch detectors and 3D-temporal models. Sequence-level defenses tailored for infrastructure surveillance, such as trajectory-based anomaly detectors and multi-camera spatiotemporal fusion systems, remain unevaluated. Future work will extend AdvSerial into adversarial training protocols, establishing it as a dynamic benchmark for validating temporal and 3D-aware security mechanisms in real infrastructure deployments.

\section{Conclusion}
\label{conclusion}

This paper presented AdvSerial, a 2D-3D joint optimization framework for generating physical adversarial attacks on high-angle pedestrian detectors deployed in infrastructure-based surveillance. Through Feature Smooth Quilting, differentiable 3D pose rendering, and Serial Frame Loss, the framework achieves durable suppression across models, views, distances, and motion states in both simulation and physical settings. AdvSerial attains an attack success rate of $89.71\%$ against YOLO-v2 in simulation and $74.8\%$ against YOLO-v5 in physical experiments, while degrading mean detection confidence from $84.30\%$ to $39.38\%$, maintaining effectiveness across eight detectors with diverse architectures and bypassing both boundary-based defenses such as NapGuard and 3D-temporal models such as Sparse4D-v3. Grad-CAM and t-SNE analyses confirm that AdvSerial operates through global semantic disruption rather than localized activation hijacking, yielding consistent confidence reduction without conspicuous artifacts and exposing temporally consistent failure modes that frame-wise or tiling-based patches fail to reveal. The findings carry direct implications for AI-based perception in transportation infrastructure, underscoring the need for defense mechanisms that incorporate motion-aware modeling, temporal attention, and explicit treatment of 3D clothing deformation. Future work will pursue online adaptation, adversarial training protocols, and hybrid optimization objectives that explicitly account for garment deformation and dynamic viewpoints, establishing AdvSerial as an evolving benchmark for assessing the security readiness of infrastructure perception systems.

\bibliographystyle{elsarticle-num-names} 
\bibliography{0article}

\end{document}